\documentclass[twoside]{article}
\usepackage[accepted]{aistats2018}
\usepackage{hyperref}       % hyperlinks
\usepackage{url}            % simple URL typesetting
\usepackage{booktabs}       % professional-quality tables
\usepackage{amsfonts}       % blackboard math symbols
\usepackage{nicefrac}       % compact symbols for 1/2, etc.
\usepackage{microtype}      % microtypography

\usepackage{amsmath, amssymb}
\usepackage{multirow}
\usepackage{algorithmic}
\usepackage{algorithm}
\usepackage{graphicx}

\usepackage{natbib}
\usepackage{mathtools, cuted}

\usepackage{multicol}

\begin{document}

\runningtitle{Mixed Membership Word Embeddings for Computational Social Science}

\twocolumn[

\aistatstitle{Mixed Membership Word Embeddings\\ for Computational Social Science}

%\aistatsauthor{ Author 1 \And Author 2 \And  Author 3 }
\aistatsauthor{ James R. Foulds }

%\aistatsaddress{ Institution 1 \And  Institution 2 \And Institution 3 } ]
\aistatsaddress{ Department of Information Systems, University of Maryland, Baltimore County.
} ]

\begin{abstract}
%  The Abstract paragraph should be indented 0.25 inch (1.5 picas) on
%  both left and right-hand margins. Use 10~point type, with a vertical
%  spacing of 11~points. The {\bf Abstract} heading must be centered,
%  bold, and in point size 12. Two line spaces precede the
%  Abstract. The Abstract must be limited to one paragraph.
Word embeddings improve the performance of NLP systems by revealing the hidden structural relationships between words.
Despite their success in many applications, word embeddings have seen very little use in computational social science NLP tasks, presumably due to their reliance on big data, and to a lack of interpretability.
I propose a probabilistic model-based word embedding method which can recover \emph{interpretable embeddings, without big data}.
The key insight is to leverage \emph{mixed membership} modeling, in which global representations are shared, but individual entities (i.e. dictionary words) are free to use these representations to uniquely differing degrees.
I show how to train the model using a combination of state-of-the-art training techniques for word embeddings and topic models.    The experimental results show an improvement in predictive language modeling of up to 63\% in MRR over the skip-gram, and demonstrate that the representations are beneficial for supervised learning. 
I illustrate the interpretability of the models with
computational social science case studies on State of the Union addresses and NIPS articles.
\end{abstract}

\section{Introduction}
\begin{table*}[t]
%\centering
\small
\begin{tabular}{cl}
\toprule
\textbf{NIPS} & reinforcement belief learning policy algorithms Singh robot machine MDP planning %algorithm  problem methods function approximation POMDP gradient markov approach based\\
\\
\midrule
\textbf{Google News} & teaching learn learning reteaching learner\_centered emergent\_literacy kinesthetic\_learning %teach learners learing lifeskills learner experiential\_learning Teaching unlearning numeracy\_literacy taught cross\_curricular Kumon\_Method ESL\_FSL\\
\\
\bottomrule
\end{tabular}
\caption{\label{tab:learningTopWords} Most similar words to ``\emph{learning},'' based on word embeddings trained on NIPS articles, and on the large generic Google News corpus \citep{mikolov2013efficient, mikolov2013distributed}.}
%\vspace{-0.5cm}
\end{table*}

Word embedding models, which learn to encode dictionary words with vector space representations,  have been shown to be valuable for a variety of natural language processing (NLP) tasks such as statistical machine translation \citep{vaswani2013decoding}, part-of-speech tagging, chunking, and named entity recogition \citep{collobert2011natural}, as they provide a more nuanced representation of words than a simple indicator vector into a dictionary.  These models follow a long line of research in data-driven semantic representations of text, including latent semantic analysis \citep{deerwester1990indexing} and its probabilistic extensions \citep{hofmann1999plsa, griffiths2007topics}.  In particular, topic models \citep{blei2003latent} have found broad applications in computational social science \citep{wallach2016CSS, roberts2014structural} and the digital humanities \citep{mimno2012computational}, where interpretable representations reveal meaningful insights.  Despite widespread success at NLP tasks, word embeddings have not yet supplanted topic models as the method of choice in computational social science applications.  I speculate that this is due to two primary factors: 1) a perceived reliance on big data, and 2) a lack of interpretability.  In this work, I develop new models to address both of these limitations.

Word embeddings have risen in popularity for NLP applications due to the success of models designed specifically for the big data setting.  In particular, \citet{mikolov2013efficient, mikolov2013distributed} showed that very simple word embedding models with high-dimensional representations can scale up to massive datasets, allowing them to outperform more sophisticated neural network language models which can process fewer documents. % given a fixed computational budget.
In this work, I offer a somewhat contrarian perspective to the currently prevailing trend of big data optimism, as exemplified by the work of \citet{mikolov2013efficient, mikolov2013distributed, collobert2011natural}, and others, who argue that massive datasets are sufficient to allow language models to automatically resolve many challenging NLP tasks.  Note that ``big'' datasets are not always available, particularly in computational social science NLP applications, where the data of interest are often not obtained from large scale sources such as the internet and social media, but from sources such as 
press releases \citep{grimmer2010bayesian}, academic journals \citep{mimno2012computational}, books \citep{zhu2015aligning}, and transcripts of recorded speech \citep{brent1999efficient, nguyen2014modeling, guo2015bayesian}.  %In this situation, a very standard practice in the literature is to train word embedding models on a separate large corpus such as Wikipedia, and use the embeddings for NLP tasks on the target smaller dataset, cf. \citep{collobert2011natural, mikolov2013efficient, pennington2014glove, vilnis2015word}.

A standard practice in the literature is to train word embedding models on a generic large corpus such as Wikipedia, and use the embeddings for NLP tasks on the target dataset, cf. \citep{collobert2011natural, mikolov2013efficient, pennington2014glove, kiros2015skip}.
However, as we shall see here, this standard practice might not always be effective, as the size of a dataset does not correspond to its degree of relevance for a particular analysis.  Even very large corpora have idiosyncrasies that can make their embeddings invalid for other domains.  For instance, suppose we would like to use word embeddings to analyze scientific articles on machine learning.  In Table \ref{tab:learningTopWords}, I report the most similar words to the word ``\emph{learning}'' based on word embedding models trained on two corpora.  For embeddings trained on articles from the NIPS conference, the most similar words are related to \emph{machine learning}, as desired, while for embeddings trained on the massive, generic Google News corpus, the most similar words relate to \emph{learning and teaching in the classroom}.  Evidently, domain-specific data can be important.

Even more concerningly, \citet{bolukbasi2016man} show that word embeddings can encode implicit sexist assumptions.  This suggests that when trained on large generic corpora they could also encode the hegemonic worldview, which is inappropriate for studying, e.g., black female hip-hop artists' lyrics, or poetry by Syrian refugees, and could potentially lead to systematic bias against minorities, women, and people of color in NLP applications with real-world consequences, such as automatic essay grading and college admissions.  In order to proactively combat these kinds of biases in large generic datasets, and to address computational social science tasks, there is a need for effective word embeddings for small datasets, so that the most relevant datasets can be used for training, even when they are small.  To make word embeddings a viable alternative to topic models for applications in the social sciences, we further desire that the embeddings are semantically meaningful to human analysts.

In this paper, I introduce an interpretable word embedding model, and an associated topic model, which are designed to work well when trained on a small to medium-sized corpus of interest. 
The primary insight is to use a data-efficient parameter sharing scheme via \emph{mixed membership} modeling, with inspiration from topic models.  Mixed membership models provide a flexible yet efficient latent representation, in which entities are associated with shared, global representations, but to uniquely varying degrees.  I identify the skip-gram word2vec model of \citet{mikolov2013efficient, mikolov2013distributed} as corresponding to a certain naive Bayes topic model, which leads to mixed membership extensions, allowing the use of \emph{fewer vectors than words}.
%In our models, the embeddings are shared between all dictionary words, allowing the use of fewer embeddings than word types, which results in a smaller latent representation to be learned while maintaining a powerful and flexible model overall.  In extensive experiments,
I show that this leads to better modeling performance without big data, as measured by predictive performance (when the context is leveraged for prediction), as well as to interpretable latent representations that are highly valuable for computational social science applications.
The interpretability of the representations arises from defining embeddings for words (and hence, documents) in terms of embeddings for topics.
%Our results indicate that our proposed models can accurately recover corpus-specific embeddings, facilitating the use of targeted representations that are relevant to a particular analysis, even when the target dataset is not ``big'' in the modern sense of the word.
My experiments also shed light on the relative merits of training embeddings on generic big data corpora versus domain-specific data.

\section{Background}
\begin{table*}[t]
%\vspace{-0.5cm}
%\hspace{-0.1cm}
\small
\centering
\begin{tabular}{ccc}
\toprule
& Skip-gram &	Skip-gram topic model\\
\midrule
\begin{minipage}[t]{1.2cm}
\ \\
Naive\\ Bayes 
\end{minipage} &
\hspace{-1cm}
\begin{minipage}[t]{8.6cm}
\begin{itemize}
\item	For each word in the corpus $w_i$
\begin{itemize}
    %\item Draw input word $w_i \sim \mbox{Discrete}(\pi)$ 
    \item For each word $w_c \in context(i)$
    \begin{itemize}
        %\item Draw $w_c | w_i$ via \\ $p(w_c|w_i) \propto exp({v'_{w_c}}^\intercal v_{w_i} + b_{w_c})$
        \item Draw $w_c$ via $p(w_c|w_i) \propto exp({v'_{w_c}}^\intercal v_{w_i} + b_{w_c})$
    \end{itemize}
\end{itemize}
\end{itemize}
\end{minipage}
&
\hspace{-1.2cm}
\begin{minipage}[t]{7.45cm}
\begin{itemize}
\item	For each word in the corpus $w_i$
\begin{itemize}
    %\item Draw input word $w_i \sim \mbox{Discrete}(\pi)$ 
    \item For each word $w_c \in context(i)$
    \begin{itemize}
        \item Draw $w_c$  via $p(w_c|w_i) = \mbox{Discrete}(\phi^{(w_i)})$
    \end{itemize}
\end{itemize}
\end{itemize}\end{minipage}
\\
\midrule
\begin{minipage}[t]{1.8cm}
\ \\
\ \\
Mixed \\ membership
\end{minipage}
 &
\hspace{-1cm}
\begin{minipage}[t]{8.6cm}
\begin{itemize}
\item	For each word in the corpus $w_i$
\begin{itemize}
    %\item Draw input word $w_i \sim \mbox{Discrete}(\pi)$ 
    \item Draw a topic $z_i \sim \mbox{Discrete}(\theta^{(w_i)})$
    \item For each word $w_c \in context(i)$
    \begin{itemize}
        \item Draw $w_c$ via $p(w_c| z_i) \propto exp({v'_{w_c}}^\intercal v_{z_i} + b_{w_c})$
    \end{itemize}
\end{itemize}
\end{itemize}
\end{minipage}
&
\hspace{-1.3cm}
\begin{minipage}[t]{7.4cm}
\begin{itemize}
\item	For each word in the corpus $w_i$
\begin{itemize}
    %\item Draw input word $w_i \sim \mbox{Discrete}(\pi)$ 
    \item Draw a topic $z_i \sim \mbox{Discrete}(\theta^{(w_i)})$
    \item For each word $w_c \in context(i)$
    \begin{itemize}
        \item Draw $w_c \mbox{ via } p(w_c| z_i)= \mbox{Discrete}(\phi^{(z_i)})$
    \end{itemize}
\end{itemize}
\end{itemize}\end{minipage} \\
\bottomrule
\end{tabular}
\caption{\label{tab:gridOfModels} ``Generative'' models.  Identifying the skip-gram (top-left)'s word distributions with topics yields analogous topic models (right), and mixed membership modeling extensions (bottom).}
\vspace{-0.5cm}
\end{table*}
In this section, I provide the necessary background on word embeddings, as well as on topic models and mixed membership models.
Traditional language models aim to predict words given the contexts that they are found in, thereby forming a joint probabilistic model for sequences of words in a language.  \citet{bengio2003neural} developed improved language models by using \emph{distributed representations} \citep{hinton1986distributed}, in which words are represented by neural network synapse weights, or equivalently, vector space embeddings. 

Later authors have noted that these \emph{word embeddings} are useful for semantic representations of words, independently of whether a full joint probabilistic language model is learned, and that alternative training schemes can be beneficial for learning the embeddings. In particular, \citet{mikolov2013efficient, mikolov2013distributed} proposed the \emph{skip-gram} model, which inverts the language model prediction task and aims to \emph{predict the context} given an input word.  
The skip-gram model is a log-bilinear discriminative probabilistic classifier parameterized by ``input'' word embedding vectors $v_{w_i}$ for the input words $w_i$, and ``output'' word embedding vectors $v'_{w_c}$ for context words $w_c \in \mbox{context}(i)$, as shown in Table \ref{tab:gridOfModels}, top-left.

Topic models such as \emph{latent Dirichlet allocation} (LDA) \citep{blei2003latent} are another class of probabilistic language models that have been used for semantic representation \citep{griffiths2007topics}.  A straightforward way to model text corpora is via unsupervised multinomial naive Bayes, in which a latent cluster assignment for each document selects a multinomial distribution over words, referred to as a \emph{topic}, with which the documents' words are assumed to be generated. LDA topic models improve over naive Bayes by using a \emph{mixed membership} model, in which the assumption that all words in a document $d$ belong to the same topic is relaxed, and replaced with a \emph{distribution} over topics $\theta^{(d)}$.  In the model's assumed generative process, for each word $i$ in document $d$, a topic assignment $z_i$ is drawn via $\theta^{(d)}$, then the word is drawn from the chosen topic $\phi^{(z_i)}$.  The mixed membership formalism provides a useful compromise between model flexibility and statistical efficiency: the $K$ topics $\phi^{(k)}$ are shared across all documents, thereby sharing statistical strength, but each document is free to use the topics to its own unique degree.  Bayesian inference further aids data efficiency, as uncertainty over $\theta^{(d)}$ can be managed for shorter documents.  Some recent papers have aimed to combine topic models and word embeddings \citep{das2015gaussian, liu2015topical}, but they do not aim to address the small data problem for computational social science, which I focus on here.
I provide a more detailed discussion of related work in the supplementary.

\section{The Mixed Membership Skip-Gram}
To design an interpretable word embedding model for small corpora, we identify novel connections between word embeddings and topic models, and adapt advances from topic modeling.  
Following the \emph{distributional hypothesis} \citep{harris1954distributional}, the skip-gram's word embeddings parameterize discrete probability distributions over words $p(w_c|w_i)$ which tend to co-occur, and tend to be semantically coherent -- a property leveraged by the Gaussian LDA model of \citet{das2015gaussian}. This suggests that these discrete distributions can be reinterpreted as \emph{topics} $\phi^{(w_i)}$.  We thus reinterpret the  skip-gram as a parameterization of a certain supervised naive Bayes topic model (Table \ref{tab:gridOfModels}, top-right).  In this topic model, input words $w_i$ are fully observed ``cluster assignments,'' and the words in $w_i$'s contexts are a ``document.''  The skip-gram differs from this supervised topic model only in the parameterization of the ``topics'' via word vectors which encode the distributions with a log-bilinear model.
Note that although the skip-gram is discriminative, in the sense that it does not jointly model the input words $w_i$, we are here equivalently interpreting it as encoding a ``conditionally generative'' process for the context given the words, in order to develop probabilistic models that extend the skip-gram.  

\begin{figure}[t]
\vspace{-0.1cm}
\includegraphics[width=\linewidth]{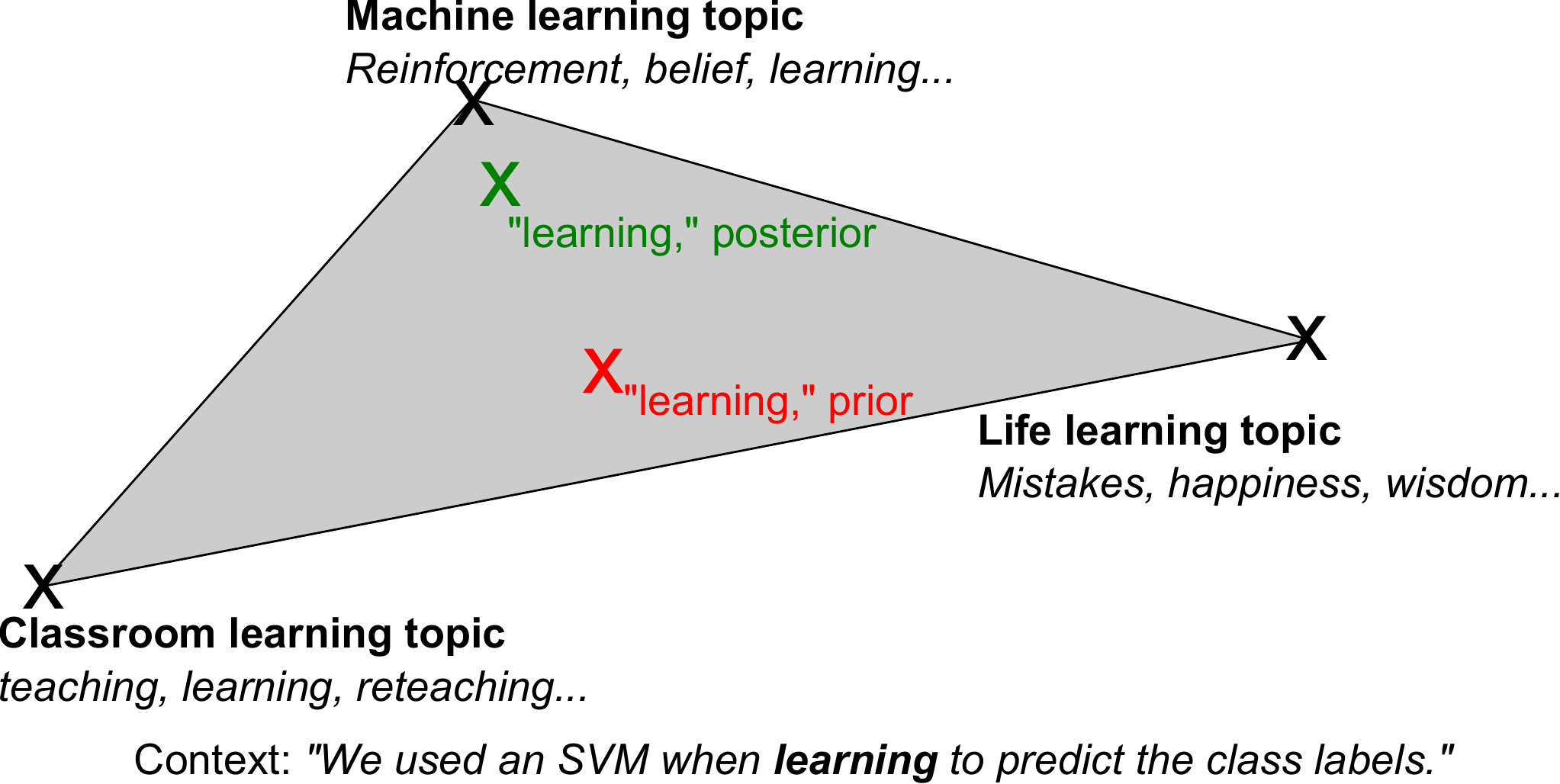}
\vspace{-0.6cm}
\caption{\label{fig:convexComb} \small %Example prior, posterior embeddings $\bar{v}_w$, $\hat{v}_{w_i}$.}
Mixed membership word embeddings $\bar{v}_w$ for word type $w$ (prior) and $\hat{v}_{w_i}$ for word token $w_i$ (posterior), are convex combinations of topic embeddings $v_k$.}
\vspace{-0.2cm}
\end{figure}

As in LDA, this model can be improved by replacing the naive Bayes assumption with a mixed membership assumption.  By applying the mixed membership representation to this topic model version of the skip-gram, we obtain the model in the bottom-right of Table \ref{tab:gridOfModels}.\footnote{The model retains a naive Bayes assumption at the context level, for latent variable count parsimony.} 
After once again parameterizing this model with word embeddings, we obtain our final model, the \emph{mixed membership skip-gram (MMSG)} (Table \ref{tab:gridOfModels}, bottom-left).  In the model, each input word has a distribution over topics $\theta^{(w)}$. Each topic has a vector-space embedding $v_k$ and each output word has a vector $v'_{w}$ (a parameter, not an embedding for $w$).  A topic $z_i \in \{1,\ldots, K\}$ is drawn for each context, and the words in the context are drawn from the log-bilinear model using $v_{z_i}$:
\begin{align}
z_i &\sim \mbox{Discrete}(\theta^{(w_i)})\\
p(w_c|z_i) &\propto exp({v'_{w_c}}^\intercal v_{z_i} + b_{w_c}) \mbox{ .}
\end{align}
We can expect that the resulting \emph{mixed membership word embeddings} are beneficial in the small-to-medium data regime for the following reasons:
\begin{enumerate}
\item By using \textbf{fewer input vectors than words}, we can reduce the size of the semantic representation to be learned (output vectors $v'_w$ are viewed as weight parameters, and not used for embedding).
\item The topic vectors are shared across all words, allowing more data to be used per vector.
\item Polysemy is addressed by clustering the words into topics, which leads to topically focused and semantically coherent vector representations.
\end{enumerate}
Of course, the model also requires some new parameters to be learned, namely the mixed membership proportions $\theta^{(w)}$.
Based on topic modeling, I hypothesized that with care, these added parameters need not adversely affect performance in the small-medium data regime, for two reasons: 1) we can use a Bayesian approach to effectively manage uncertainty in them, and to marginalize them out, which prevents them being a bottleneck during training; and 2) at test time, using the posterior for $z_i$ given the context, instead of the ``prior'' $p(z_i|w_i,\theta) = \theta^{(w_i)}$, mitigates the impact of uncertainty in $\theta^{(w_i)}$ due to limited training data: %.  By Bayes' rule, we have:
\begin{align}
p(z_{i} &= k|w_i, \mbox{context}(i), \mathbf{V}, \mathbf{V}', \mathbf{b}, \theta)  \label{eqn:topicInference} \\ & \propto \theta^{(w_i)}_k \prod_{c \in \mbox{context}(i)} \frac{exp( v_{w_c}^{\prime\intercal} v_{k} + b_{w_c})}{\sum_{j'=1}^Vexp( v^{\prime \intercal}_{j'} v_{k} + b_{j'})} \mbox{ .} \nonumber
\end{align}
To obtain a vector for a word type $w$, we can use the prior mean, $\bar{v}_{w} \triangleq \sum_k v_k\theta^{(w)}_k$.  For a word token $w_i$, we can leverage its context via the posterior mean, $ \hat{v}_{w_i} \triangleq \sum_k v_k p(z_{i} = k|w_i, \mbox{context}(i), \mathbf{V}, \mathbf{V}', \mathbf{b}, \theta)$.  These embeddings are convex combinations of topic vectors (see Figure \ref{fig:convexComb} for an example).
With fewer vectors than words, some model capacity is lost, but the flexibility of the mixed membership representation allows the model to compensate.  When the number of shared vectors equals the number of words, the mixed membership skip-gram is strictly more representationally powerful than the skip-gram.  With more vectors than words, we can expect that the increased representational power would be beneficial in the big data regime. As this is not my goal, I leave this for future work.  
\section{Training Algorithm for the MMSG}
\setlength{\textfloatsep}{10pt}
\begin{algorithm}[t]
\begin{algorithmic}
\FOR{$j=1:\mbox{maxAnnealingIter}$}
		\STATE $T_j := T_0 + \lambda \kappa^j$
		\FOR{$i=1:\mbox{N}$}
			\STATE $c \sim \mbox{Uniform}(|\mbox{context}(w_i)|)$;
			\STATE  $z_i^{(new)} \sim q_{w_c}$; //via cached  alias table samples
			%\STATE $z_i^{(new)} \sim q_{w_c^{(i)}}$ //using cached samples from alias tables, done in amortized $O(1)$ time
			\STATE accept or reject $z_i^{(new)}$ via Equation \ref{eqn:acceptRatio};
			\STATE If accept, $z_i := z_i^{(new)}$;
		\ENDFOR
\ENDFOR
\STATE $\hat{\theta}^{(w_i)}_k :\propto n_k^{(w_i)\neg i} + \alpha_k$
\STATE $[\mathbf{V}, \mathbf{V}', b] := \mbox{NCE}(inputWords = \mathbf{z},$ \STATE $\ \ \ \ \ \ \ \ \ \ \ \ \ \ \ \ \ \ \ \ \ \ \ \ contextWords = \mathbf{w})$;
\end{algorithmic}
\caption{Training the mixed membership skip-gram via annealed MHW and NCE \label{alg:trainingMMWE}}
\end{algorithm}
\setlength{\textfloatsep}{18pt}
I first describe an idealized but impractical training algorithm for the MMSG, and then introduce a more practicable procedure (Algorithm \ref{alg:trainingMMWE}).  The MMSG can in principle be trained via maximum likelihood estimation using EM.  Optimizing the log-likelihood is hindered by the latent variables, which EM circumvents by focusing on the complete-data log-likelihood (CDLL), $\log p(\mathbf{w}, \mathbf{z}|\mathbf{V}, \mathbf{V}', \mathbf{b}, \theta ) =$
\begin{align}
&\sum_{i=1}^N \sum_{k=1}^K z_{i,k} \log \theta_k^{(w_i)} \ + \sum_{i=1}^N \sum_{k=1}^K z_{i,k}  \times \label{eqn:CDLL} \\
&\sum_{c \in \mbox{context}(i)} \Big (v^{\prime \intercal}_{w_c} v_{k} + b_{w_c} -\log {\sum_{j'=1}^Dexp( v^{\prime \intercal}_{j'} v_{k} + b_{j'})} \Big ) \mbox{ .} \nonumber
\end{align}
%The E-step computes the expected complete data log-likelihood by way of ``E-step responsibilities'' $\gamma_{i,k}$:
The E-step computes the \emph{E-step responsibilities} $\gamma_{i,k}$:
\begin{align*}
%\gamma_{i,k} = E[z_{i,k}|\mathbf{V}^{(old), \theta^{(old)}}, w_i, \mbox{context}(i)] = Pr(z_{i_k} = 1|\mathbf{V}^{(old)}, w_i, \mbox{context}(i), \theta^{(old)})
%\gamma_{i,k} = Pr(z_{i_k} = 1|\mathbf{V}^{(old)},\mathbf{V}^{\prime (old)}, \mathbf{b}^{(old)}, w_i, \mbox{context}(i), \theta^{(old)}) \mbox{ ,}
\gamma_{i,k} = p(z_{i_k} = 1|w_i, \mbox{context}(i), \{\mathbf{V},\mathbf{V}', \mathbf{b}, \theta \}^{(old)}) \mbox{ ,}
\end{align*}
where $(old)$ superscripts denote current parameter estimates.  The M-step optimizes the lower bound on the log-likelihood obtained by substituting $\gamma_{i,k}$ for $z_{i,k}$ in Equation \ref{eqn:CDLL}.  However, this involves a $O(KD)$ complexity for both the E- and M-steps for each token, where $K$ and $D$ are the number of topics/dictionary words, respectively, and even $O(D)$ per token is considered impractical for word embeddings \citep{mnih2012fast, mikolov2013efficient}.  %Note that \cite{tian2014probabilistic}'s model assumes a small number of prototypes per word (e.g. they have $K \leq 10$), while in our case we would like the number of topics $K$ to generally be in the thousands or tens of thousands.
Instead, I propose an approximation to EM that is \emph{sublinear time} in both $K$ and $D$.  We first impute $\mathbf{z}$ using a reparameterization technique, thereby reducing the task to standard word embedding.  This can be done in sublinear time using the  Metropolis-Hastings-Walker (MHW) algorithm.  With an oracle $\hat{\mathbf{z}}$ for $\mathbf{z}$, the log-likelihood $\log p(\mathbf{w} |\mathbf{V},\mathbf{V}', \mathbf{b}, \theta ) = \log \sum_{\mathbf{z}} p(\mathbf{w}, \mathbf{z}| \mathbf{V},\mathbf{V}', \mathbf{b}, \theta ) $ simplifies to the CDLL $\log p(\mathbf{w}, \hat{\mathbf{z}}| \mathbf{V}, \mathbf{V}', \mathbf{b}, \theta )$, as in Equation \ref{eqn:CDLL}.
%This approximates the log-likelihood objective function with the CDLL, plugging in a best guess for $\mathbf{z}$, in a manner reminiscent of the hard EM algorithm.
%
%Alternatively, by EM's Jensen's inequality argument, $\log Pr(\mathbf{w}, \hat{\mathbf{z}}| \mathbf{V}, \mathbf{V}', \mathbf{b}, \theta )$ lower bounds $\log Pr(\mathbf{w} |\mathbf{V}, \mathbf{V}', \mathbf{b}, \theta )$, for any  $\hat{\mathbf{z}}$. %This actually follows trivially from the sum rule of probability (see above), and is a terrible bound. It's better not to use that argument, which is useless, although correct.
%Given $\hat{\mathbf{z}}$,
We then efficiently learn the topic and word embeddings via noise-contrastive estimation (NCE). With enough computation NCE exactly optimizes our CDLL objective function, but avoids computing expensive normalization constants and provides an adjustable computational efficiency knob.  The details are described below.
%The overall algorithm is summarized in Algorithm \ref{alg:trainingMMWE}.

%\subsection{Approximate E-step: Topic Model Pre-Clustering}
%\subsection{Imputing the $\mathbf{z}$'s: Topic Model Pre-Clustering via Annealed Metropolis-Hastings-Walker}
\subsection{Imputing the $\mathbf{z}$'s}
\begin{table*}[t]
\small
%\hspace{-1cm}
\centering
\begin{tabular}{ll}
\toprule
\multicolumn{2}{c}{Input word = ``Bayesian''}\\
Model & Top words in topic for input word. Top 3 topics for word shown for mixed membership models. \\
%\midrule
\midrule
SGTM &  model networks learning neural bayesian data models approach network framework \\%prior algorithm methods mackay distribution probability inference gaussian set function \\
SG & belief learning framework models methods markov function bayesian based inference\\
&\\
%MMSGTM & bayesian model parameters posterior prior distribution approach likelihood variational inference\\
%&    neural networks computation bayesian learning mackay framework network functions practical \\
%&    carlo monte bayesian gaussian neural neal implementation methods models williams \\
MMSGTM & neural bayesian networks mackay computation framework practical learning weigend backpropagation\\
&    model models bayesian    prior    data    parameters    likelihood    priors    structure    graphical \\
&     monte    carlo    chain    markov    sampling    mcmc    method    methods    model    bayesian \\
%MMSG &  variational likelihood bayesian inference approach parameters marginal dirichlet posterior sampling\\
%& neural bayesian learning networks computation framework regularization entropy press mackay\\
%&   neal rasmussen monte bayesian models http press neural barber carlo\\
MMSG &  neural networks weigend bayesian    data    mackay    learning    computation    practical   \\% vol\\
& probability    model    data    models    priors    algorithm    bayesian    likelihood    set    parameters\\
&   carlo  monte    mcmc    chain    reversible    sampling    model    posterior   \\% neural    data\\
\bottomrule
\end{tabular}
\caption{\label{tab:exampleTopics}
SG = skip-gram, TM = topic model, MM = mixed membership. }
\vspace{-0.7cm}
\end{table*}

To derive such an algorithm, the key insight is that our MMSG model (Table \ref{tab:gridOfModels}, bottom left)  is equivalent to the topic model version (Table \ref{tab:gridOfModels}, bottom right), up to the parameterization.  With sufficiently high dimensional embeddings, the log-bilinear model can capture any distribution $p(w_c| z_i)$, and so the maximum likelihood embeddings would encode the exact same word distributions as the MLE topics for the topic model, $\phi^{(z_i)}$.  However, the topic model admits a collapsed Gibbs sampler (CGS) that efficiently resolves the cluster assignments, which cause the bottleneck during EM.   I therefore propose to reparameterize the MMSG as its corresponding topic model for the purposes of imputing the $\mathbf{z}$'s. %, and run simulated annealing based on the collapsed Gibbs sampler for the topic model to impute the topic assignments. % (i.e. to cluster the words).  %This \emph{pre-clustering} step solves the expensive E-step for the remainder of the training procedure.  We can think of this as both learning the cluster assignments and fixing the $\theta$ and $\phi$ parameters implied by these assignments.  
Then, with the $\mathbf{z}$'s fixed to the estimate $\hat{\mathbf{z}}$, learning the word and topic vectors corresponds to finding the optimal vectors for encoding the $\phi$'s. % via the log-bilinear model.

This topic model pre-clustering step is reminiscent of \citet{reisinger2010multi,huang2012improving, liu2015topical}, who apply an off-the-shelf clustering algorithm (or LDA) to initially identify different clusters of contexts, and then apply word embedding algorithms on the cluster assignments.  However, our clustering is learned based on the word embedding model itself, and clustering at test time is performed via Bayesian reasoning, in Equation \ref{eqn:topicInference}, rather than via an ad-hoc method.  With Dirichlet priors on the parameters, the collapsed Gibbs update is (derivation in the supplement):
\begin{align}
p(z_i &= k|\cdot) \propto \Big ( n_k^{(w_i)\neg i} + \alpha_k \Big ) \\
&\times \prod_{c=1}^{|\mbox{context}(i)|} \frac{n_{w_c}^{(k)\neg_i} + \beta_{w_c} + n_{w_c}^{(i,c)}}{n^{(k)\neg i} + \sum_{w'}\beta_{w'} + c - 1} \mbox{ ,}  \nonumber
\end{align}
where $\alpha$ and $\beta$ are parameter vectors for Dirichlet priors over the topic and word distributions, $n_k^{(w_i)}$ and $n_{w_c}^{(k)\neg_i}$ are input and output word/topic counts (excluding the current word), and $n_{w_c}^{(i,c)}$ is the number of occurrences of word $w_c$ before the $c$th word in the $i$th context.  %By using modern sparse implementation techniques, the collapsed Gibbs sampler can be made to scale sublinearly in $K$ \citep{yao2009efficient, li2014reducing}.
We scale this algorithm up to thousands of topics using an adapted version of the recently proposed Metropolis-Hastings-Walker algorithm for high-dimensional topic models, which scales sublinearly in $K$ \citep{li2014reducing}. The method uses a data structure called an \emph{alias table}, which allows for amortized O(1) time sampling from discrete distributions.  A Metropolis-Hastings update is used to correct for approximating the CGS update with a proposal distribution based on these samples.
%  This in turn facilitates an efficient implementation of the topic model. The alias table samples from slightly stale distributions, and a Metropolis-Hastings update is used to correct for this.
%In the Metropolis-Hastings-Walker algorithm, the Gibbs update is split into a mixture over a sparse component, and a dense-slow-changing component, in our case:
%\begin{align}
%p(z_i = k|\cdot) \propto n_k^{(w_i)\neg i} A_{ik} + \alpha_kA_{ik} & \mbox{ , } & 
% A_{ik} = \prod_{c=1}^{|\mbox{context}(i)|} \frac{n_{w^{(i)}_c}^{(k)\neg_i} + \beta_{w_c^{(i)} + n_{w_c^{(i)}}^{(i,c)}}}{n^{(k)\neg i} + \sum_{w'}\beta_{w'} + c - 1} \mbox{ .}
%\end{align}
%We approximate the second, dense term via samples from the alias tables.
We can interpret the product over the context, which dominates the collapsed Gibbs update, as a \emph{product of experts} \citep{hinton2002training}, where each word in the context is an “expert” which weighs in multiplicatively on the update.
%In order to approximate this via alias tables, we use a Metropolis proposal which approximates the product of experts with a “mixture of experts.” In the proposal, a random word is drawn from the context, and the topic is drawn proportional to its contribution to the update. Since the dense terms (from the context) are expected to dominate the update, we do not use the sparse term in the proposal, unlike \citet{li2014reducing}, so sampling from the proposal is amortized constant time, rather than linear in the sparsity pattern.
%We expect this proposal to have some resemblance to the target distribution, but be somewhat flatter, which is a property we’d generally like in a proposal distribution:
%\begin{align}
% q(k) = \sum_{c=1}^{|\mbox{context}(w_i)|} \frac{1}{|\mbox{context}|}q_{w_c^{(i)}}(k) \mbox{ , }  q_{w_c^{(i)}}(k) = \frac{1}{Z_{w_c}} \alpha_k \frac{n_{w^{(i)}_c}^{(k)} + \beta_{w_c^{(i)}}}{n^{(k)} + \sum_{w'}\beta_{w'}} \mbox{ .}
%\end{align}
%\begin{align}
% q(k) = \sum_{c=1}^{|\mbox{context}(w_i)|} \frac{1}{|\mbox{context}|}q_{w_c^{(i)}}(k) \mbox{ , }  q_{w_c^{(i)}}(k) \propto \frac{n_{w^{(i)}_c}^{(k)} + \beta_{w_c^{(i)}}}{n^{(k)} + \sum_{w'}\beta_{w'}} \mbox{ .}
%\end{align}
In order to approximate this via alias tables, we use  proposals which approximate the product of experts with a “mixture of experts.” We select a word $w_c$ uniformly from the context, and the proposal $q_{w_c}$ draws a candidate topic proportionally to the chosen context word's contribution to the update:
\begin{align}
 c \sim \mbox{Uniform}(|\mbox{context}(w_i)|) \mbox{ , } 
 q_{w_c}(k) \propto \frac{n_{w_c}^{(k)} + \beta_{w_c}}{n^{(k)} + \sum_{w'}\beta_{w'}} \mbox{ .} \nonumber %I want to number this equation, but it falls to the next line. Include number in the unlikely event that there is space
\end{align}
 %Since the dense terms (from the context) are expected to dominate the update, we do not use the sparse term in the proposal, unlike \citet{li2014reducing}, so
We can expect these proposals to bear a resemblance to the target distribution, but to be flatter, which is a property we'd generally like in a proposal distribution. The proposal is implemented efficiently by sampling from the experts via the alias table data structure, in amortized O(1) time, rather than in time linear in the sparsity pattern, as in \citep{li2014reducing}, since the proposal does not involve the sparse term (which is less important in our case).  We perform simulated annealing to optimize over the posterior, which is very natural for Metropolis-Hastings.  Interpreting the negative log posterior as the energy function for a Boltzmann distribution at temperature $T_j$ for iteration $j$, this is achieved by raising the model part of the Metropolis-Hastings acceptance ratio to the power of $\frac{1}{T_j}$:
%\begin{align}
%z_i^{(new)} \sim q \mbox{\ , \ \ \ \ } 
%Pr(\mbox{accept } z_i^{(new)}| \cdot ) = \min \Big ( 1,  \big (\frac{p(z_i = z_i^{(new)}| \cdot )}{p(z_i = z_i^{(old)}| \cdot )} \frac{q(z_i^{(old)})}{q(z_i^{(new)})} \big )^{\frac{1}{T_j}} \Big) \mbox{ .}
%\end{align}
\begin{align}
 & z_i^{(new)}\sim q_{w_c} \mbox{ , }
 p(\mbox{accept } z_i^{(new)}| \cdot ) = \nonumber \\
& \ \ \ \min \Big ( 1,  \big (\frac{p(z_i = z_i^{(new)}| \cdot )}{p(z_i = z_i^{(old)}| \cdot )}  \big )^{\frac{1}{T_j}} \frac{q_{w_c}(z_i^{(old)})}{q_{w_c}(z_i^{(new)})} \Big) \mbox{ .} \label{eqn:acceptRatio}
\end{align}
%%%%%%%%Derivation for the above%%%%%%%%
%\begin{align*}
%\frac{\exp( \frac{\log p(z^{new})}{T} )}{\exp( \frac{\log p(z^{old})}{T} )} \frac{q(z_i^{(old)})}{q(z_i^{(new)})} &= \exp( \frac{\log p(z^{new}) - \log p(z^{old})}{T}) \frac{q(z_i^{(old)})}{q(z_i^{(new)})} \\
%&= \exp( \frac{1}{T}\log \frac{p(z^{new})}{p(z^{old})}) \frac{q(z_i^{(old)})}{q(z_i^{(new)})}\\
%&= \exp ( \log  \big ( \Big (\frac{p(z^{new})}{p(z^{old})} \Big )^{\frac{1}{T}} \big ) ) \frac{q(z_i^{(old)})}{q(z_i^{(new)})}\\
%&=  \big (\frac{p(z^{new})}{p(z^{old})} \big)^{\frac{1}{T}} \frac{q(z_i^{(old)})}{q(z_i^{(new)})}\\
%\end{align*}
%%%%%%%%End derivation%%%%%%%%
Annealing also helps with mixing, as the standard Gibbs updates can become nearly deterministic.  We use a temperature schedule $T_j = T_0 + \lambda \kappa^j$, where $T_0$ is the target final temperature, $\kappa < 1$, and $\lambda$ controls the initial temperature, and therefore mixing in the early iterations.  In my experiments, I use $T_0 = 0.0001$, $\kappa = 0.99$, and $\lambda = |\mbox{context}|$. The acceptance probability can be computed in time constant in $K$, and sampling is amortized constant time in $K$, so each iteration is in amortized constant time in $K$.  %After running the annealed collapsed Gibbs sampler for a sufficient number of iterations, sampled $\mathbf{z}$ assignments and
Rao-Blackwellized estimates of the mixed membership proportions are obtained from the final sample as $\hat{\theta}^{(w_i)}_k \propto n_k^{(w_i)\neg i} + \alpha_k$.

\subsection{Learning the Embeddings}
Finally, with the topic assignments $\hat{\mathbf{z}}$ imputed and $\theta$ estimated via the topic model, we must learn the embeddings, which is still an expensive $O(D)$ per context for maximum likelihood estimation, i.e. optimizing
\begin{align}
%\hspace{-2.2mm}
\log p(\mathbf{w}, \hat{\mathbf{z}}| \vec{\mathbf{V}}, \mathbf{b}, \theta ) = \log p(\mathbf{w} |\hat{\mathbf{z}}, \vec{\mathbf{V}}, \mathbf{b})+ \mbox{const,} \label{eqn:MLEwithOracle}
\end{align}
where $\vec{\mathbf{V}}$ is the vector of all word and topic embeddings.
This same complexity is also an issue for the standard skip-gram, which \citet{mnih2012fast, mnih2013learning} have addressed using the noise-contrastive estimation (NCE) algorithm of \citet{gutmann2010noise, gutmann2012noise}.  NCE avoids the expensive normalization step, making the algorithm scale sublinearly in the vocabulary size $D$.
The algorithm solves unsupervised learning tasks by transforming them into the supervised learning task of distinguishing the data from randomly sampled noise samples, via logistic regression.
% The algorithm solves unsupervised learning tasks by transforming them into the supervised learning task of distinguishing the data from randomly sampled noise.
%NCE takes as input the log-likelihood of a data point, i.e. a context word given an input word and its topic assignment
%\begin{align}
%\log Pr(w_c| \vec{\mathbf{V}}, w_i, z_i, \mathbf{b}) &= \log Pr^0(w_c| \vec{\mathbf{V}}, w_i, z_i, \mathbf{b}) +  a \nonumber \\
%&= v^{\prime \intercal}_{w_{c}} v_{z_i} + b_{w_c} + a \mbox{ ,}
%\end{align}
% where $Pr^0$ refers to an unnormalized distribution, and $a$ is a parameter encoding the corresponding log normalization constant in the current context $i$. Following \citet{mnih2012fast}, we fix $a=0$, under the supposition that the NCE procedure will compensate for this by encouraging the distributions to ``self-normalize'' in order to optimize the NCE objective.
%NCE performs logistic regression to distinguish between the data samples and the noise samples.  
Supposing that there are $k$ samples from the noise distribution per word-pair example, the NCE objective function for context $i$ is
 \begin{align}
 \hspace{-0.1cm}
 J^{(i)}(\vec{\mathbf{V}}, \mathbf{b}) &\triangleq E_{p^{(i)}_d}[\log \sigma(G(w_c; \vec{\mathbf{V}}, w_i, z_i, \mathbf{b}))] \nonumber 
 \\ &- kE_{p_n}[\log (1 - \sigma(G(w_c; \vec{\mathbf{V}}, w_i, z_i, \mathbf{b})))] \label{eqn:NCEperContext}
 \end{align}
 where $p^{(i)}_d$ is the data distribution for words $w_c$ context $i$, and $G(w_c; \vec{\mathbf{V}}, w_i, z_i, \mathbf{b}) \triangleq \log p( w_c| \vec{\mathbf{V}}, w_i, z_i, \mathbf{b}) - \log p_n( w_c)$ is the difference in log-likelihood between the model and the noise distributions.
% The stochastic gradient to the overall objective based on the $c$th word in the $i$th context and $k$ corresponding noise samples ${w_c^{(i)}}^{(j)}$ is
% \begin{align}
% \frac{\partial}{\partial \vec{\mathbf{V}}}J^{(i,c)}(\vec{\mathbf{V}}) =& (1 - \sigma(G(w_c^{(i)}; \vec{\mathbf{V}}, w_i, z_i)))\frac{\partial}{\partial \vec{\mathbf{V}}}\log Pr(w_c^{(i)}| \vec{\mathbf{V}}, w_i, z_i) \nonumber \\
% -& \sum_{j=1}^k \Big [\sigma(G({w_c^{(i)}}^{(j)}; \vec{\mathbf{V}}, w_i, z_i)) \frac{\partial}{\partial \vec{\mathbf{V}}}\log Pr( {w_c^{(i)}}^{(j)}| \vec{\mathbf{V}}, w_i, z_i) \Big ] \mbox{ .}
% \end{align}
 We learn the embeddings by stochastic gradient ascent on the NCE objective.   As the number of noise samples tends to infinity, the method increasingly well approximates maximum likelihood estimation, i.e. the stationary points of Equation \ref{eqn:NCEperContext} converge on those of Equation \ref{eqn:MLEwithOracle} \citep{gutmann2010noise, gutmann2012noise}. %, while avoiding explicitly computing the normalization constant or its derivative, as required by a direct optimization of the log-likelihood.

\section{Experimental Results}
\label{sec:experiments}
The goals of our experiments were to study the relative merits of big data and domain-specific small data, to validate the proposed methods, and to study their applicability for computational social science research.

\subsection{Quantitative Experiments}

\begin{table*}[t]
%\small
%\footnotesize
\scriptsize
%\hspace{-1.1cm}
\hspace{-0.33cm}
\centering
\begin{tabular}{ccccccccccc}
\toprule
Dataset & Frequency  & Google   & SG & SG       & MMSG & MMSG      & SGTM & SGTM     & MMSGTM  & MMSGTM\\
        & baseline   & +context &    & +context & prior& posterior &      & +context & prior   & posterior \\
\midrule
NIPS & 0.029 & 0.027 & 0.038 & 0.031 & 0.037 & \textbf{0.062} & \textbf{0.055} & \textbf{0.064} & \textbf{0.046} & \textbf{0.074}  \\
SOTU & 0.021 &  0.021  & 0.025 & 0.023 & 0.022 &  \textbf{0.034} &  \textbf{0.036} & \textbf{0.046} & \textbf{0.032} & \textbf{0.045}\\
Shakespeare & 0.015 & \textbf{0.032} & 0.020 &  0.010  & 0.015 & 0.019 & \textbf{0.025} & \textbf{0.043} & 0.020 & \textbf{0.025} \\ %100 topics. The MMSG posterior and MMSGTM prior have no significant difference from SG (so the MMSG posterior's loss is not significant, and so everything from MMSG posterior onwards never loses vs the skip-gram)
Du Bois & 0.028 & 0.033  & 0.045 & 0.037 & 0.041 & \textbf{0.053} & \textbf{0.052} & \textbf{0.081} &  \textbf{0.050} & \textbf{0.066}  \\
%Federalist\\
\bottomrule      
\end{tabular}
\caption{\label{tab:MRR} Mean reciprocal rank of held-out context words. SG = skip-gram, TM = topic model, MM = mixed membership. Bold indicates statistically significant improvement versus SG. %, at the 5\% level.
}
\end{table*}

\begin{table*}[t]
%\small
%\footnotesize
\scriptsize
%\hspace{-1.1cm}
%\hspace{-0.33cm}
\centering
\begin{tabular}{cccccccccc}
\toprule
Dataset         & \#Classes & \#Topics & Tf-idf  & Google & MMSG & SG & MMSGTM & SG+MMSG &  SG+MMSG+Google \\ % & All \\
\midrule 
20 Newsgroups     & 20 & 200 & 83.33 & 52.50 & 55.58 & 59.50 & 64.08 & 66.55 & 72.53 \\ %& 83.18 (83.175 before rounding)\\
Reuters-150       & 150 & 500 & 73.04 & 53.65  & 65.26 & 69.53 & 66.97 & 70.63 & 71.20 \\%& 73.14 \\
Ohsumed   & 23 & 500 & 43.07 & 20.56 & 31.82 & 37.57 & 32.41 & 39.53 & 40.27 \\ %& 43.12\\ %Note: 20K version
\midrule
SOTU (RMSE) & Regression & 500 & 19.57 & 8.64 & 12.73 & 10.57 & 21.88
 & 9.94  & 8.15  \\%&\\
\bottomrule      
\end{tabular}
\caption{\label{tab:classification} Document categorization (top, classification accuracy, larger is better), and predicting the year of State of the Union addresses (bottom, RMSE, LOO cross-validation, smaller is better). %Bold indicates statistically significant improvement versus SG.
}
\end{table*}

\begin{figure*}[t]
\vspace{-0.2cm}
\includegraphics[width=\linewidth]{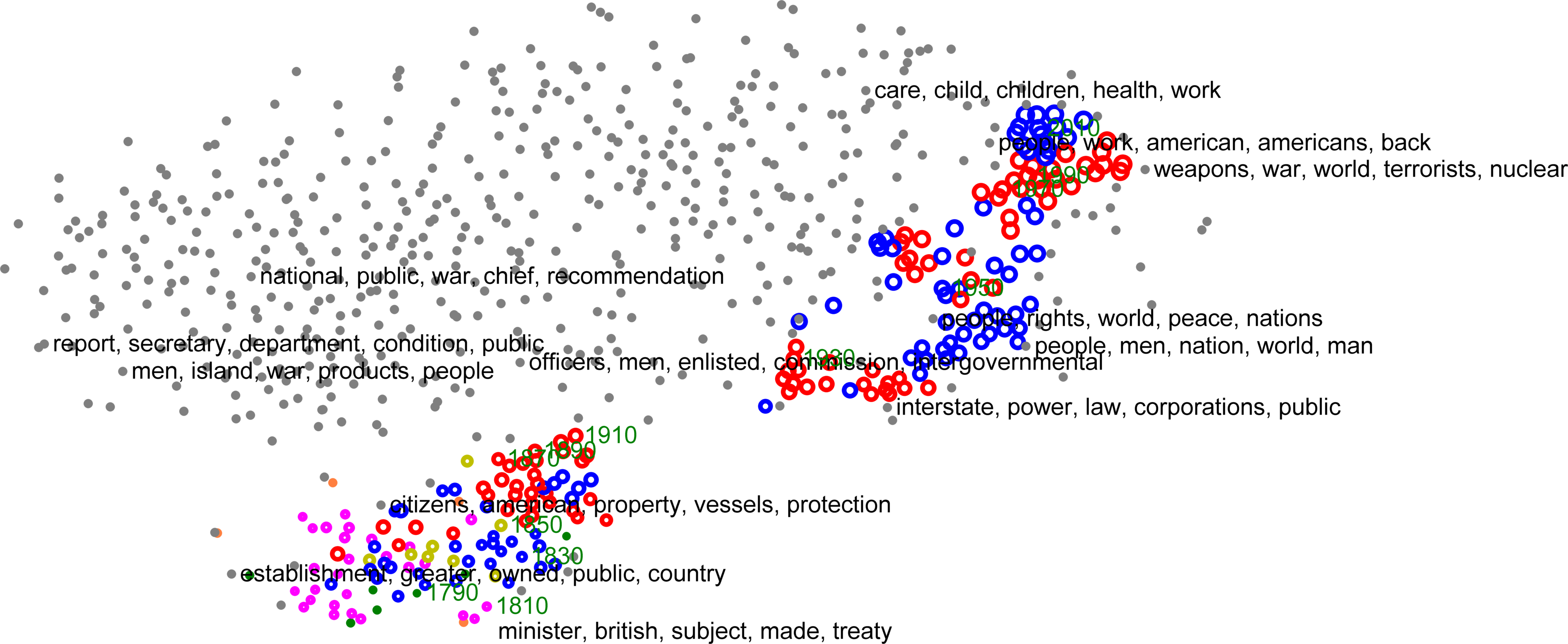}
\vspace{-0.2cm}
\caption{\label{fig:SOTU} \small State of the Union (SOTU) addresses. Colored circles are $t$-SNE projected embeddings for SOTU addresses.  Color = party (red = GOP, blue = Democrats, light green = Whigs, pink = Democratic-Republicans, orange = Federalists (John Adams), green = George Washington), size = recency (year, see dates in green). Gray circles correspond to topics.}
\end{figure*}
\begin{figure*}
%\hspace{0.5cm}
\vspace{-0.3cm}

\begin{minipage}{0.71\linewidth}
\small
%\hspace{1cm}
\begin{tabular}{ll}
\toprule
\multicolumn{2}{c}{Nearest topic after composition of mean vectors for words}\\ %(normalized prior mean vectors)}\\
\midrule
object + recognition &
objects visual object recognition model\\
character + recognition & recognition segmentation character\\ %handwriting \\%network\\
speech + recognition &  speech recognition hmm system hybrid\\
%    'model'  'markov'  'hidden'    'continuous'    'neural'
computer + vision & computer vision ieee image pattern\\
computer + science & university science colorado department\\ %computer\\
bias + variance & error training set data performance\\
covariance + variance &  gaussian distribution model matrix \\%covariance
\bottomrule
%\multicolumn{2}{c}{Nearest topics after vector composition}\\
\end{tabular}
\hspace{-3cm}
\end{minipage}
\hspace{-2cm}
\begin{minipage}{0.3\linewidth}
%\hspace{-1cm}
\includegraphics[width=1.2\linewidth]{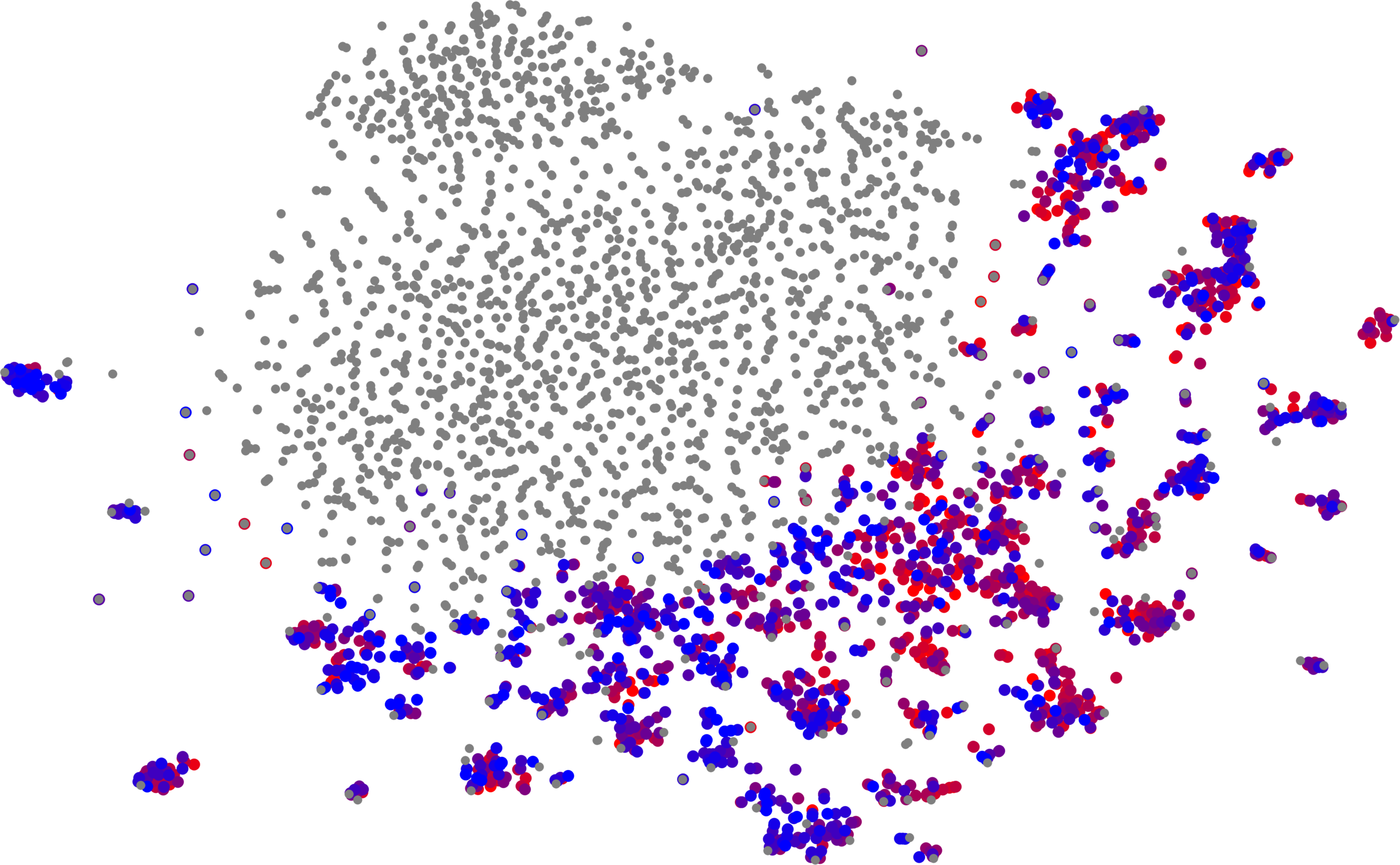}
%\vspace{0.5cm}
\end{minipage}
%\noindent\makebox[\linewidth]{\rule{\linewidth}{0.4pt}}
\caption{\label{fig:composition} \textbf{Left:} Vector compositionality examples, NIPS. \textbf{Right:}  NIPS documents/ topics, $t$-SNE.}
\vspace{-0.3cm}
\end{figure*}

I first measured the effectiveness of the embeddings at the \emph{skip-gram's training task}, predicting context words $w_c$ given input words $w_i$.  This task measures the methods' performance for predictive language modeling.  I used four datasets of sociopolitical, scientific, and literary interest: the corpus of NIPS articles from 1987 -- 1999 ($N \approx 2.3$ million), the U.S. presidential state of the Union addresses from 1790 -- 2015 ($N \approx 700,000$), the complete works of Shakespeare ($N \approx 240,000$; this version did not contain the Sonnets), and the writings of black scholar and activist W.E.B. Du Bois, as digitized by Project Gutenberg ($N \approx 170,000$).  For each dataset, I held out 10,000 $(w_c, w_i)$ pairs uniformly at random, where $w_c \in \mbox{context}(i), |\mbox{context}(i)| = 10$, and aimed to predict $w_c$ given $w_i$ (and optionally, $\mbox{context}(i) \setminus w_c$).  Since there are a large number of classes, I treat this as a ranking problem, and report the mean reciprocal rank.  The experiments were repeated and averaged over 5 train/test splits.

The results are shown in Table \ref{tab:MRR}.  I compared to a word frequency baseline, the skip-gram (SG), and Tomas Mikolov/Google's vectors trained on Google News, $N\approx 100$ billion, via CBOW.  Simulated annealing was performed for 1,000 iterations, NCE was performed for 1 million minibatches of size 128, and 128-dimensional embeddings were used (300 for Google). %(the naive Bayes topic model is fully supervised and requires no iteration).
I used $K=2,000$ for NIPS, $K=500$ for state of the Union, and $K=100$ for the two smaller datasets. Methods were able to leverage the remainder of the context, either by adding the context's vectors, or via the posterior (Equation \ref{eqn:topicInference}), which helped for all methods except the naive skip-gram. We can identify several noteworthy findings.  First, the generic big data vectors (Google+context) were outperformed by the skip-gram on 3 out of 4 datasets (and by the skip-gram topic model on the other), by a large margin, indicating that domain-specific embeddings are often important.  Second, the mixed membership models, using posterior inference, beat or matched their naive Bayes counterparts, for both the word embedding models and the topic models.  As hypothesized, posterior inference on $z_i$ at test time was important for good performance. 
Finally, the topic models beat their corresponding word embedding models at prediction. 
I therefore recommend the use of our MMSG topic model variant for predictive language modeling in the small data regime.

\subsubsection{Downstream Tasks}
I tested the performance of the representations as features for document categorization and regression tasks.  The results are given in Table \ref{tab:classification}.  For document categorization, I used three standard benchmark datasets: \emph{20 Newsgroups} (19,997 newsgroup posts),  \emph{Reuters-150} newswire articles 
(15,500 articles and 150 classes), and \emph{Ohsumed} medical abstracts on 23 cardiovascular diseases (20,000 articles).\footnote{All document categorization datasets were obtained from \url{http://disi.unitn.it/moschitti/corpora.htm}.}  I held out 4,000 test documents for 20 Newsgroups, and used the standard train/test splits from the literature in the other corpora (e.g. for \emph{Ohsumed}, 50\% of documents were assigned to training and to test sets).
I obtained document embeddings for the MMSG, in the same latent space as the topic embeddings, by summing the posterior mean vectors  $\hat{v}_{w_i}$ for each token.  Vector addition was similarly used to construct document vectors for the other embedding models.  All vectors were normalized to unit length.  I also considered a tf-idf baseline.  Logistic regression models were trained on the features extracted on the training set for each method.

Across the three datasets, several clear trends emerged (Table \ref{tab:classification}).  First, the generic Google vectors were consistently and substantially outperformed in classification performance by the skipgram (SG) and MMSG vectors, highlighting the importance of corpus-specific embeddings.  Second, despite the MMSG's superior performance at language modeling on small datasets, the SG features outperformed the MMSG's at the document categorization task.  By encoding vectors at the topic level instead of the word level, the MMSG loses word level resolution in the embeddings, which turned out to be valuable for these particular classification tasks.  We are not, however, restricted to use only one type of embedding to construct features for classification.  Interestingly, when the SG and MMSG features were concatenated (SG+MMSG), this improved classification performance over these vectors individually.  This suggests that the topic-level MMSG vectors and word-level SG vectors encode \emph{complementary} information, and both are beneficial for performance.  Finally, further concatenating the generic Google vectors' features (SG+MMSG+Google) improved performance again, despite the fact that these vectors performed poorly on their own.  It should be noted that tf-idf, which is notoriously effective for document categorization, outperformed the embedding methods on these datasets.

I also analyzed the regression task of predicting the year of a state of the Union address based on its text information.  I used lasso-regularized linear regression models, evaluated via a leave-one-out cross-validation experimental setup.  Root-mean-square error (RMSE) results are reported in Table \ref{tab:classification} (bottom).  Unlike for the other tasks, the Google big data vectors were the best individual features in this case, outperforming the domain-specific SG and MMSG embeddings individually. %, perhaps due to the small size of the dataset.
%The concatenation of the MMSG vectors with the SG vectors once again improved over the individual embeddings' performance.
On the other hand, SG+MMSG+Google performed the best overall, showing that domain-specific embeddings can improve performance even when big data embeddings are successful.  The tf-idf baseline  %performed poorly on this task, and
was beaten by all of the embedding models on this task.

\subsection{Computational Social Science Case Studies: State of the Union and NIPS}
I also performed several case studies.  I obtained document embeddings, in the same latent space as the topic embeddings, by summing the posterior mean vectors  $\hat{v}_{w_i}$ for each token, %(and similarly for author embeddings), 
and visualized them in two dimensions using $t$-SNE \citep{maaten2008visualizing} (all vectors were normalized to unit length).  The state of the Union addresses (Figure \ref{fig:SOTU}) are embedded almost linearly by year, with a major jump around the New Deal (1930s), and are well separated by party at any given time period. The embedded topics (gray) allow us to interpret the space. The George W. Bush addresses are embedded near a ``war on terror'' topic (``weapons, war...''), and the Barack Obama addresses are embedded near a ``stimulus'' topic (``people, work...'').

On the NIPS corpus, for the input word ``Bayesian'' (Table \ref{tab:exampleTopics}), the naive Bayes and skip-gram models learned a topic with words that refer to Bayesian networks, probabilistic models, and neural networks.  %The mixed membership models are able to separate this into more coherent and specific topics including Bayesian inference, Bayesian training of neural networks (for which Sir David MacKay was a strong proponent), and Monte Carlo methods (championed by Radford Neal, as well as Carl Rasmussen, Chris Williams, and David Barber, in the context of Gaussian Processes).
The mixed membership models are able to separate this into more coherent and specific topics including Bayesian modeling, Bayesian training of neural networks (for which Sir David MacKay was a strong proponent, and Andreas Weigend wrote an influential early paper), and Monte Carlo methods.
By performing the additive composition of word vectors, which we obtain by finding the prior mean vector for each word type $w$, $\bar{v}_{w} \triangleq \sum_k v_k\theta^{(w)}_k$ (and then normalizing), we obtain relevant topics $v_k$ as nearest neighbors (Figure \ref{fig:composition}).  %This can allow an analyst to rapidly find topics of interest based on keywords.
Similarly, we find that the additive composition of topic and word vectors works correctly: $v_{objectRecognition} - \bar{v}_{object} + \bar{v}_{speech} \approx v_{speechRecognition}$, and $v_{speechRecognition} - \bar{v}_{speech} + \bar{v}_{character} \approx v_{characterRecognition} $.

The $t$-SNE visualization of NIPS documents (Figure \ref{fig:composition}) shows some temporal clustering patterns (blue documents are more recent, red documents are older, and gray points are topics).  I provide a more detailed case study on NIPS 
%, including higher resolution visualizations of NIPS documents and authors, and a discussion of these results,
in the supplementary material.

\section{Conclusion}
I have proposed a model-based method for training interpretable corpus-specific word embeddings for computational social science, using mixed membership representations, Metropolis-Hastings-Walker sampling, and NCE.  Experimental results for prediction, supervised learning, and case studies on state of the Union addresses and NIPS articles, indicate that high-quality embeddings and topics can be obtained using the method.
The results highlight the fact that big data is not always best, as domain-specific data can be very valuable, even when it is small.
I plan to use this approach for substantive social science applications, and to address algorithmic bias and fairness issues.

%\subsubsection*{Acknowledgements}
\textbf{Acknowledgements}\\ %Technically I should use subsubsection* but that pushes it to a new page, and this looks identical
I thank Eric Nalisnick and Padhraic Smyth for many helpful discussions.
%We also plan to extend our models to automatically leverage big data sets together with a target small data set, in order to get the best of both.
%
%From AISTATS 2017 template
%\subsubsection*{Acknowledgements}
%
%Use unnumbered third level headings for the acknowledgements.  All
%acknowledgements go at the end of the paper.
%
%\textbf{Acknowledgments:} 

%\subsubsection*{References}

%\medskip
\small

\bibliographystyle{apalike}
\bibliography{references}

\begin{thebibliography}{}

\bibitem[Airoldi et~al., 2008]{Airoldi2008}
Airoldi, E., Blei, D., Feinberg, S., and Xing, E. (2008).
\newblock {Mixed membership stochastic blockmodels}.
\newblock {\em Journal of Machine Learning Research}, 9:1981--2014.

\bibitem[Airoldi et~al., 2014]{airoldi2014introduction}
Airoldi, E.~M., Blei, D.~M., Erosheva, E.~A., and Fienberg, S.~E. (2014).
\newblock Introduction to mixed membership models and methods.
\newblock In {\em Handbook of Mixed Membership Models and Their Applications}.
  Chapman and Hall/CRC.

\bibitem[Barnard et~al., 2003]{barnard2003matching}
Barnard, K., Duygulu, P., Forsyth, D., Freitas, N.~d., Blei, D.~M., and Jordan,
  M.~I. (2003).
\newblock Matching words and pictures.
\newblock {\em Journal of Machine Learning Research}, 3(Feb):1107--1135.

\bibitem[Bengio et~al., 2003]{bengio2003neural}
Bengio, Y., Ducharme, R., Vincent, P., and Jauvin, C. (2003).
\newblock A neural probabilistic language model.
\newblock {\em Journal of Machine Learning Research}, 3:1137--1155.

\bibitem[Blei et~al., 2003]{blei2003latent}
Blei, D., Ng, A., and Jordan, M. (2003).
\newblock Latent {D}irichlet allocation.
\newblock {\em Journal of Machine Learning Research}, 3:993--1022.

\bibitem[Bolukbasi et~al., 2016]{bolukbasi2016man}
Bolukbasi, T., Chang, K.-W., Zou, J.~Y., Saligrama, V., and Kalai, A.~T.
  (2016).
\newblock Man is to computer programmer as woman is to homemaker? {D}ebiasing
  word embeddings.
\newblock In {\em Advances in Neural Information Processing Systems 29}, pages
  4349--4357.

\bibitem[Brent, 1999]{brent1999efficient}
Brent, M.~R. (1999).
\newblock An efficient, probabilistically sound algorithm for segmentation and
  word discovery.
\newblock {\em Machine Learning}, 34(1):71--105.

\bibitem[Collobert et~al., 2011]{collobert2011natural}
Collobert, R., Weston, J., Bottou, L., Karlen, M., Kavukcuoglu, K., and Kuksa,
  P. (2011).
\newblock Natural language processing (almost) from scratch.
\newblock {\em Journal of Machine Learning Research}, 12(Aug):2493--2537.

\bibitem[Das et~al., 2015]{das2015gaussian}
Das, R., Zaheer, M., and Dyer, C. (2015).
\newblock Gaussian {LDA} for topic models with word embeddings.
\newblock In {\em Proceedings of the 53nd Annual Meeting of the Association for
  Computational Linguistics}, pages 795--804.

\bibitem[Deerwester et~al., 1990]{deerwester1990indexing}
Deerwester, S., Dumais, S.~T., Furnas, G.~W., Landauer, T.~K., and Harshman, R.
  (1990).
\newblock Indexing by latent semantic analysis.
\newblock {\em Journal of the American Society for Information Science},
  41(6):391.

\bibitem[Erosheva et~al., 2004]{erosheva2004mixed}
Erosheva, E., Fienberg, S., and Lafferty, J. (2004).
\newblock Mixed-membership models of scientific publications.
\newblock {\em Proceedings of the National Academy of Sciences of the United
  States of America}, 101(Suppl 1):5220--5227.

\bibitem[Erosheva, 2003]{erosheva2003bayesian}
Erosheva, E.~A. (2003).
\newblock Bayesian estimation of the grade of membership model.
\newblock {\em Bayesian Statistics}, 7:501--510.

\bibitem[Fei-Fei and Perona, 2005]{fei2005bayesian}
Fei-Fei, L. and Perona, P. (2005).
\newblock A {B}ayesian hierarchical model for learning natural scene
  categories.
\newblock In {\em Proceedings of the 2005 IEEE Conference on Computer Vision
  and Pattern Recognition (CVPR)}, pages 524--531. IEEE.

\bibitem[Griffiths et~al., 2007]{griffiths2007topics}
Griffiths, T.~L., Steyvers, M., and Tenenbaum, J.~B. (2007).
\newblock Topics in semantic representation.
\newblock {\em Psychological Review}, 114(2):211.

\bibitem[Grimmer, 2010]{grimmer2010bayesian}
Grimmer, J. (2010).
\newblock A {B}ayesian hierarchical topic model for political texts: Measuring
  expressed agendas in senate press releases.
\newblock {\em Political Analysis}, pages 1--35.

\bibitem[Guo et~al., 2015]{guo2015bayesian}
Guo, F., Blundell, C., Wallach, H., and Heller, K. (2015).
\newblock The {B}ayesian echo chamber: Modeling social influence via linguistic
  accommodation.
\newblock In {\em Proceedings of the International Conference on Artificial
  Intelligence and Statistics (AISTATS)}, pages 315--323.

\bibitem[Gutmann and Hyv{\"a}rinen, 2010]{gutmann2010noise}
Gutmann, M. and Hyv{\"a}rinen, A. (2010).
\newblock Noise-contrastive estimation: A new estimation principle for
  unnormalized statistical models.
\newblock In {\em Proceedings of the International Conference on Artificial
  Intelligence and Statistics (AISTATS)}.

\bibitem[Gutmann and Hyv{\"a}rinen, 2012]{gutmann2012noise}
Gutmann, M.~U. and Hyv{\"a}rinen, A. (2012).
\newblock Noise-contrastive estimation of unnormalized statistical models, with
  applications to natural image statistics.
\newblock {\em Journal of Machine Learning Research}, 13(Feb):307--361.

\bibitem[Harris, 1954]{harris1954distributional}
Harris, Z.~S. (1954).
\newblock Distributional structure.
\newblock {\em Word}, 10(2-3):146--162.

\bibitem[Hinton, 2002]{hinton2002training}
Hinton, G.~E. (2002).
\newblock Training products of experts by minimizing contrastive divergence.
\newblock {\em Neural Computation}, 14(8):1771--1800.

\bibitem[Hinton et~al., 1986]{hinton1986distributed}
Hinton, G.~E., Mcclelland, J.~L., and Rumelhart, D.~E. (1986).
\newblock Distributed representations.
\newblock In {\em Parallel Distributed Processing: Explorations in the
  Microstructure of Cognition. Volume 1: Foundations}, chapter~3, pages
  77--109. MIT Press, Cambridge, MA.

\bibitem[Hinton and Salakhutdinov, 2009]{hinton2009replicated}
Hinton, G.~E. and Salakhutdinov, R.~R. (2009).
\newblock Replicated softmax: an undirected topic model.
\newblock In {\em Advances in Neural Information Processing Systems}, pages
  1607--1614.

\bibitem[Hofmann, 1999a]{hofmann1999plsa}
Hofmann, T. (1999a).
\newblock Probabilistic latent semantic analysis.
\newblock In {\em Proceedings of the Fifteenth International Conference on
  Uncertainty in Artificial Intelligence}, pages 289--296. Morgan Kaufmann
  Publishers Inc.

\bibitem[Hofmann, 1999b]{hofmann1999plsi}
Hofmann, T. (1999b).
\newblock Probabilistic latent semantic indexing.
\newblock In {\em Proceedings of the 22nd Annual International ACM SIGIR
  Conference on Research and Development in Information Retrieval}, pages
  50--57. ACM.

\bibitem[Huang et~al., 2012]{huang2012improving}
Huang, E.~H., Socher, R., Manning, C.~D., and Ng, A.~Y. (2012).
\newblock Improving word representations via global context and multiple word
  prototypes.
\newblock In {\em Proceedings of the 50th Annual Meeting of the Association for
  Computational Linguistics: Long Papers-Volume 1}, pages 873--882. Association
  for Computational Linguistics.

\bibitem[Kiros et~al., 2015]{kiros2015skip}
Kiros, R., Zhu, Y., Salakhutdinov, R.~R., Zemel, R., Urtasun, R., Torralba, A.,
  and Fidler, S. (2015).
\newblock Skip-thought vectors.
\newblock In {\em Advances in Neural Information Processing Systems}, pages
  3294--3302.

\bibitem[Larochelle and Lauly, 2012]{NIPS2012_4613}
Larochelle, H. and Lauly, S. (2012).
\newblock A neural autoregressive topic model.
\newblock In Pereira, F., Burges, C. J.~C., Bottou, L., and Weinberger, K.~Q.,
  editors, {\em Advances in Neural Information Processing Systems 25}, pages
  2708--2716.

\bibitem[Li et~al., 2014]{li2014reducing}
Li, A.~Q., Ahmed, A., Ravi, S., and Smola, A.~J. (2014).
\newblock Reducing the sampling complexity of topic models.
\newblock In {\em Proceedings of the 20th ACM SIGKDD International Conference
  on Knowledge Discovery and Data Mining}, pages 891--900. ACM.

\bibitem[Liu et~al., 2015]{liu2015topical}
Liu, Y., Liu, Z., Chua, T.-S., and Sun, M. (2015).
\newblock Topical word embeddings.
\newblock In {\em Proceedings of the AAAI Conference on Artificial
  Intelligence}, pages 2418--2424.

\bibitem[Maaten and Hinton, 2008]{maaten2008visualizing}
Maaten, L. v.~d. and Hinton, G. (2008).
\newblock Visualizing data using t-{SNE}.
\newblock {\em Journal of Machine Learning Research}, 9(Nov):2579--2605.

\bibitem[Manton et~al., 1994]{manton1994statistical}
Manton, K.~G., Tolley, H.~D., and Woodbury, M.~A. (1994).
\newblock {\em Statistical applications using fuzzy sets}.
\newblock Wiley-Interscience.

\bibitem[Mikolov et~al., 2013a]{mikolov2013efficient}
Mikolov, T., Chen, K., Corrado, G., and Dean, J. (2013a).
\newblock Efficient estimation of word representations in vector space.
\newblock {\em Proceedings of the 2013 International Conference on Learning
  Representations (ICLR)}.

\bibitem[Mikolov et~al., 2013b]{mikolov2013distributed}
Mikolov, T., Sutskever, I., Chen, K., Corrado, G., and Dean, J. (2013b).
\newblock Distributed representations of words and phrases and their
  compositionality.
\newblock In {\em Advances in Neural Information Processing Systems}, pages
  3111--3119.

\bibitem[Mimno, 2012]{mimno2012computational}
Mimno, D. (2012).
\newblock Computational historiography: Data mining in a century of classics
  journals.
\newblock {\em Journal on Computing and Cultural Heritage (JOCCH)}, 5(1):3.

\bibitem[Mnih and Kavukcuoglu, 2013]{mnih2013learning}
Mnih, A. and Kavukcuoglu, K. (2013).
\newblock Learning word embeddings efficiently with noise-contrastive
  estimation.
\newblock In {\em Advances in Neural Information Processing Systems}, pages
  2265--2273.

\bibitem[Mnih and Teh, 2012]{mnih2012fast}
Mnih, A. and Teh, Y.~W. (2012).
\newblock A fast and simple algorithm for training neural probabilistic
  language models.
\newblock In {\em Proceedings of the 29th International Conference on Machine
  Learning}.

\bibitem[Nguyen et~al., 2014]{nguyen2014modeling}
Nguyen, V.-A., Boyd-Graber, J., Resnik, P., Cai, D.~A., Midberry, J.~E., and
  Wang, Y. (2014).
\newblock Modeling topic control to detect influence in conversations using
  nonparametric topic models.
\newblock {\em Machine Learning}, 95(3):381--421.

\bibitem[Pennington et~al., 2014]{pennington2014glove}
Pennington, J., Socher, R., and Manning, C.~D. (2014).
\newblock Glove: Global vectors for word representation.
\newblock In {\em Proceedings of the 2014 Conference on Empirical Methods in
  Natural Language Processing (EMNLP)}, volume~14, pages 1532--1543.

\bibitem[Pritchard et~al., 2000]{pritchard2000inference}
Pritchard, J.~K., Stephens, M., and Donnelly, P. (2000).
\newblock Inference of population structure using multilocus genotype data.
\newblock {\em Genetics}, 155(2):945--959.

\bibitem[Reisinger and Mooney, 2010]{reisinger2010multi}
Reisinger, J. and Mooney, R.~J. (2010).
\newblock Multi-prototype vector-space models of word meaning.
\newblock In {\em Human Language Technologies: The 2010 Annual Conference of
  the North American Chapter of the Association for Computational Linguistics},
  pages 109--117. Association for Computational Linguistics.

\bibitem[Roberts et~al., 2014]{roberts2014structural}
Roberts, M.~E., Stewart, B.~M., Tingley, D., Lucas, C., Leder-Luis, J.,
  Gadarian, S.~K., Albertson, B., and Rand, D.~G. (2014).
\newblock Structural topic models for open-ended survey responses.
\newblock {\em American Journal of Political Science}, 58(4):1064--1082.

\bibitem[Teh et~al., 2006]{teh2006hierarchical}
Teh, Y.~W., Jordan, M.~I., Beal, M.~J., and Blei, D.~M. (2006).
\newblock Hierarchical {D}irichlet processes.
\newblock {\em Journal of the American Statistical Association}, 101(476).

\bibitem[Tian et~al., 2014]{tian2014probabilistic}
Tian, F., Dai, H., Bian, J., Gao, B., Zhang, R., Chen, E., and Liu, T.-Y.
  (2014).
\newblock A probabilistic model for learning multi-prototype word embeddings.
\newblock In {\em Proceedings of the International Conference on Computational
  Linguistics (COLING)}, pages 151--160.

\bibitem[Vaswani et~al., 2013]{vaswani2013decoding}
Vaswani, A., Zhao, Y., Fossum, V., and Chiang, D. (2013).
\newblock Decoding with large-scale neural language models improves
  translation.
\newblock In {\em Proceedings of the 2013 Conference on Empirical Methods in
  Natural Language Processing (EMNLP)}, pages 1387--1392.

\bibitem[Wallach, 2016]{wallach2016CSS}
Wallach, H. (2016).
\newblock Computational social science: Toward a collaborative future.
\newblock In Alvarez, R.~M., editor, {\em Computational Social Science:
  Discovery and Prediction}. Cambridge University Press.

\bibitem[Zhu et~al., 2015]{zhu2015aligning}
Zhu, Y., Kiros, R., Zemel, R., Salakhutdinov, R., Urtasun, R., Torralba, A.,
  and Fidler, S. (2015).
\newblock Aligning books and movies: Towards story-like visual explanations by
  watching movies and reading books.
\newblock In {\em Proceedings of the IEEE International Conference on Computer
  Vision}, pages 19--27.

\end{thebibliography}

% % % % % Supplementary material!!
\clearpage 
\appendix
\twocolumn[
  \begin{@twocolumnfalse}
    \part*{Supplementary Material}
  \end{@twocolumnfalse}
  ]
\section{Related Work}
In this supplementary document, we discuss related work in the literature and its relation to our proposed methods, provide a case study on NIPS articles, and derive the collapsed Gibbs sampling update for the MMSGTM, which we leverage when training the MMSG.
\subsection{Topic Modeling and Word Embeddings}
The \emph{Gaussian LDA} model of \citet{das2015gaussian} improves the performance of topic modeling by leveraging the semantic information encoded in word embeddings.  Gaussian LDA modifies the generative process of LDA such that each topic is assumed to generate the vectors via its own Gaussian distribution.  Similarly to our MMSG model, in Gaussian LDA each topic is encoded with a vector, in this case the mean of the Gaussian.  It takes pre-trained word embeddings as input, rather than learning the embeddings from data within the same model, and does not aim to perform word embedding.

The topical word embedding (TWE) models of \citet{liu2015topical} reverse this, as they take LDA topic assignments of words as input, and aim to use them to improve the resultant word embeddings.  The authors propose three variants, each of which modifies the skip-gram training objective to use LDA topic assignments together with words.  In the best performing variant, called \emph{TWE-1}, a standard skip-gram word embedding model is trained independently with another skip-gram variant, which tries to predict context words given the input word's topic assignment.  The skip-gram embedding and the topic embeddings are concatenated to form the final embedding.

At test time, a distribution over topics for the word given the context, $p(z_i|\mbox{context}(i))$ is estimated according to the topic counts over the other context words.  Using this as a prior, a posterior over topics given both the input word and the context is calculated, and similarities between pairs of words (with their contexts) are averaged over this posterior, in a procedure inspired by those used by \citet{reisinger2010multi, huang2012improving}.  The primary similarity to our MMSG approach is the use of a training algorithm involving the prediction of context words, given a topic.  Our method does this as part of an overall model-based inference procedure, and we learn mixed membership proportions $\theta^{(w)}$ rather than using empirical counts as the prior over topics for a word token.  In accordance with the skip-gram's prediction model, we are thus able to model the context words in the data likelihood term when computing the posterior probability of the topic assignment. TWE-1 requires that topic assignments are available at test time.  It provides a mechanism to predict contextual similarity, but not to predict held-out context words, so we are unable to compare to it in our experiments.  

Other neurally-inspired topic models include replicated softmax \citep{hinton2009replicated}, and its successor, DocNADE \citep{NIPS2012_4613}.  Replicated softmax extends the restricted Boltzmann machine to handle multinomial counts for document modeling.  DocNADE builds on the ideas of replicated softmax, but uses the NADE architecture, where observations (i.e. words) are modeled sequentially given the previous observations.

\begin{figure*}[t]
\begin{minipage}{\linewidth}
\includegraphics[width=\linewidth]{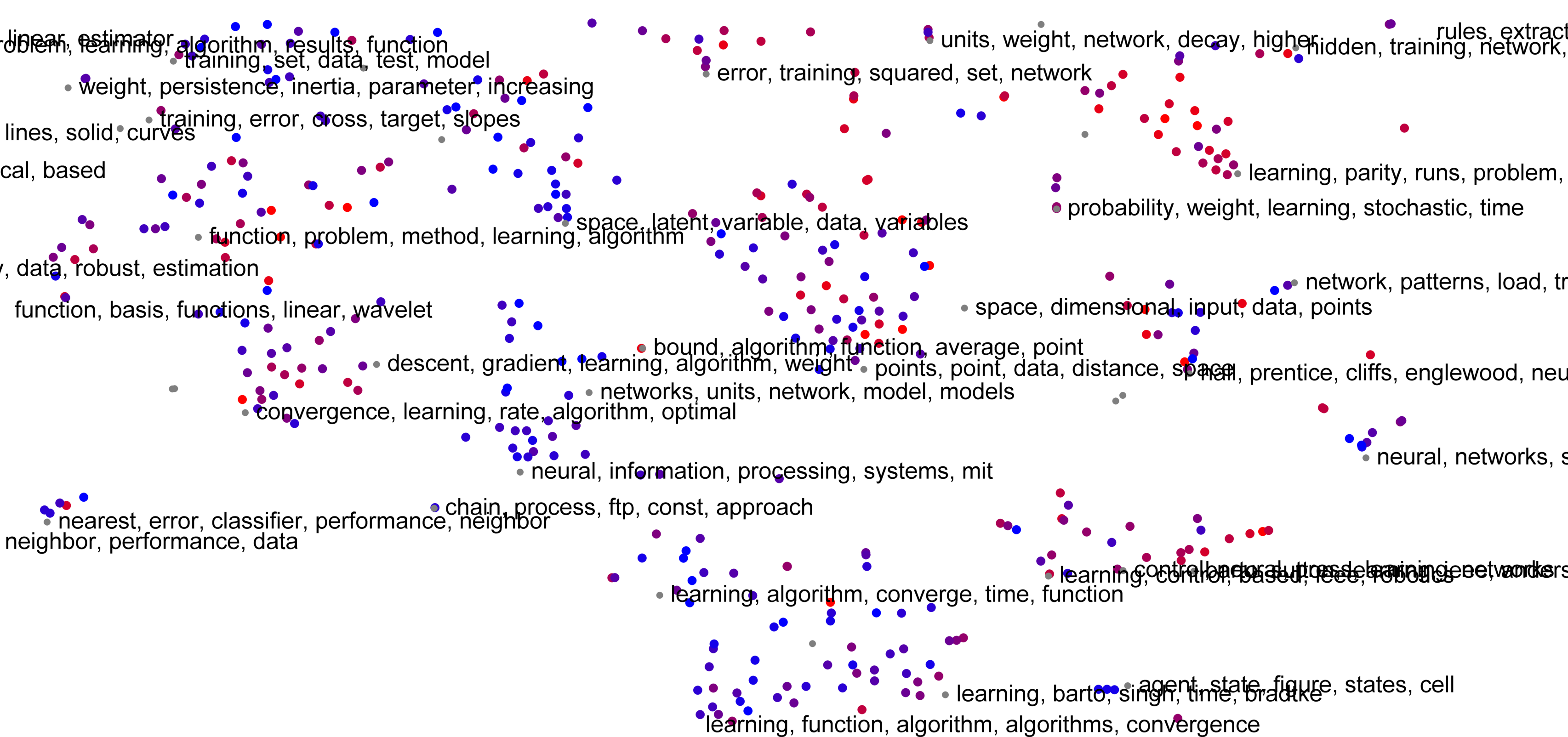}
%\vspace{0.1cm}
\end{minipage}
% % % Puts full visualization and closest topics for author at bottom
%\begin{minipage}{0.3\linewidth}
%\includegraphics[width=1.2\linewidth]{figures/NIPS_2.png}
%\end{minipage}
%\begin{minipage}{0.75\linewidth}
%\small
%\hspace{1cm}
%\begin{tabular}{ll}
%\toprule
%T Sejnowski & learning, algorithm, problem, model, training\\
%C Koch & circuit, response, figure, high, frequency\\
%M Jordan & model, probability, state, algorithm, data \\
%G Hinton & input, features, network, information, system\\
%M Mozer& input, output, unit, network, units\\
%%P Dayan & learning, model, performance, network, results\\
%%S Singh & learning, function, algorithm, algorithms, convergence\\
%%Y Bengio & input, features, network, information, system\\
%%J Moody & error, training, set, data, performance\\
%%V Tresp & training, set, data, test, model\\
%%C Williams & data, model, figure, points, components\\
%%A Barto & learning, barto, singh, time, bradtke\\
%Z Ghahramani & model, data, parameters, network, training\\
%%R Lippmann & nearest, classifiers, neighbor, performance, data\\
%J Platt & training, data, number, set, network\\
%C Bishop & space, latent, variable, data, variables\\
%S Amari & sources, separation, signals, source, algorithm\\
%P Bartlett & case, functions, function, linear, estimator\\
%\bottomrule
%\end{tabular}
%\end{minipage}
%\caption{\label{fig:NIPSdocs}$t$-SNE visualization of NIPS documents/ topics (top: zoomed in, bottom-left: full visualization). Blue/red = more recent/older, gray = topics. Closest topic per author, NIPS (bottom-right).}
\caption{\label{fig:NIPSdocs} NIPS documents/topics, $t$-SNE, zoomed in. Blue/red = more recent/older, gray = topics.}
\vspace{-0.3cm}
\end{figure*}

\begin{figure*}[t]
%\vspace{-1.7cm}
\includegraphics[width=\linewidth]{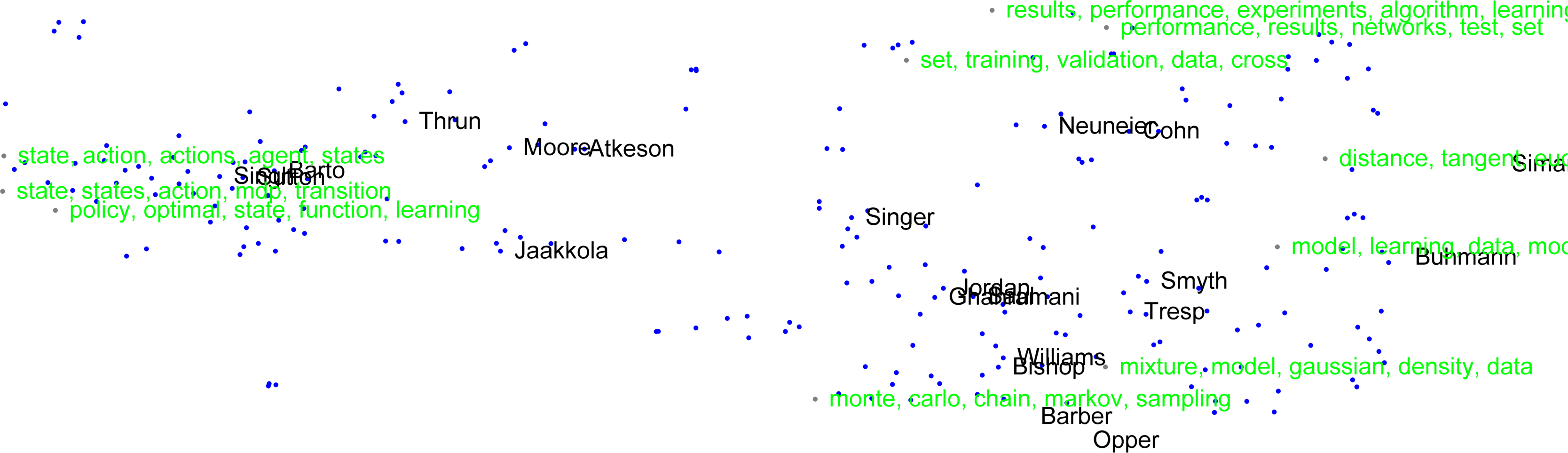}
\vspace{-0.3cm}
\caption{\label{fig:NIPSauthors} NIPS authors and topics, $t$-SNE, zoomed in. Blue = authors, gray = topics.}
\end{figure*}
\subsection{Multi-Prototype Embedding Models}
Multi-prototype embeddings models are another relevant line of work.  These models address lexical ambiguity by assigning multiple vectors to each word type, each corresponding to a different meaning of that word.  \citet{reisinger2010multi} propose to cluster the occurrences of each word type, based on features extracted from its context.  Embeddings are then learned for each cluster.  \citet{huang2012improving} apply a similar approach, but they use initial single-prototype word embeddings to provide the features used for clustering.  These clustering methods have some resemblance to our topic model pre-clustering step, although their clustering is applied within instances of a given word type, rather than globally across all word types, as in our methods.  This results in models with more vectors than words, while we aim to find fewer vectors than words, to reduce the model's complexity for small datasets.  Rather than employing an off-the-shelf clustering algorithm and then applying an unrelated embedding model to its output, our approach aims to perform model-based clustering within an overall joint model of topic/cluster assignments and word vectors.

Perhaps the most similar model to ours in the literature is the probabilistic multi-prototype embedding model of \citet{tian2014probabilistic}, who treat the prototype assignment of a word as a latent variable, assumed drawn from a mixture over prototypes for each word.  The embeddings are then trained using EM.  Our MMSG model can be understood as the mixed membership version of this model, in which the prototypes (vectors) are shared across all word types, and each word type has its own mixed membership proportions across the shared prototypes.  While a similar EM algorithm can be applied to the MMSG, the E-step is much more expensive, as we typically desire many more shared vectors (often in the thousands) than we would prototypes per a single word type (Tian et al. use ten in their experiments).  We use the Metropolis-Hastings-Walker algorithm with the topic model reparameterization of our model in order to address this by efficiently pre-solving the E-step.

\subsection{Mixed Membership Modeling}

Mixed membership modeling is a flexible alternative to traditional clustering, in which each data point is assigned to a single cluster. Instead, mixed membership models posit that individual entities are associated with multiple underlying clusters, to differing degrees, as encoded by a mixed membership vector that sums to one across the clusters \citep{erosheva2004mixed, airoldi2014introduction}.  These mixed membership proportions are generally used to model lower-level grouped data, such as the words inside a document.  Each lower-level data point inside a group is assumed to be assigned to one of the shared, global clusters according to the group-level membership proportions.  Thus, a mixed membership model consists of a mixture model for each group, which share common mixture component parameters, but with differing mixture proportions.

This formalism has lead to probabilistic models for a variety of applications, including medical diagnosis \citep{manton1994statistical}, population genetics \citep{pritchard2000inference}, survey analysis \citep{erosheva2003bayesian}, computer vision \citep{barnard2003matching, fei2005bayesian}, text documents \citep{hofmann1999plsi, blei2003latent}, and social network analysis \citep{Airoldi2008}.  Nonparametric Bayesian extensions, in which the number of underlying clusters is learned from data via Bayesian inference, have also been proposed \citep{teh2006hierarchical}.  In this work, dictionary words are assigned a mixed membership distribution over a set of shared latent vector space embeddings.  Each instantiation of a dictionary word (an ``input'' word) is assigned to one of the shared embeddings based on its dictionary word's membership vector.  The words in its context (``output'' words) are assumed to be drawn based on the chosen embedding.
%\newpage
\section{Case Study on NIPS}

In Figure \ref{fig:NIPSdocs}, we show a zoomed in $t$-SNE visualization of NIPS document embeddings.  We can see regions of the space corresponding to learning algorithms (bottom), data space and latent space (center), training neural networks (top), and nearest neighbors (bottom-left).  We also visualized the authors' embeddings via $t$-SNE (Figure \ref{fig:NIPSauthors}). We find regions of latent space for reinforcement learning authors (left: ``state, action,...,'' Singh, Barto,Sutton), probabilistic methods (right: ``mixture, model,'' ``monte, carlo,'' Bishop, Williams, Barber, Opper, Jordan, Ghahramani, Tresp, Smyth), and evaluation (top-right: ``results, performance, experiments,...'').  %The closest topics per author are intuitive (Figure \ref{fig:NIPSdocs}, bottom-right).

\section{Derivation of the Collapsed Gibbs Update}
Let $C_i = |\mbox{context}(i)|$ be the number of output words in the $i$th context, let $w_1^{(i)}, \ldots, w_{C_i}^{(i)}$ be those output words, and let $\mathbf{w}_{\neg i}$ be the input words other that $w_i$ (similarly, topic assignments $\mathbf{z}_{\neg i}$ and output words $\mathbf{w}^{(\neg i)} $).  Then the collapsed Gibbs update samples from the conditional distribution
\begin{align*}
p(z_i &= k|\mathbf{z}_{\neg i},w_i, w^{(i)}_1, \ldots, w^{(i)}_{C_{i}}, \mathbf{w}_{\neg i}, \mathbf{w}^{(\neg i)}, \alpha, \beta)\\
\propto\ & p(z_i = k, w^{(i)}_1, \ldots, w^{(i)}_{C_i}|\mathbf{z}_{\neg i}, w_i, \mathbf{w}_{\neg i}, \mathbf{w}^{(\neg i)}, \alpha, \beta)\\
&= \int_{\phi^{(k)}} \int_{\theta^{(w_i)}} p(z_i = k, w^{(i)}_1, \ldots, w^{(i)}_{C_i}, \phi^{(k)}, \theta^{(w_i)}|\mathbf{z}_{\neg i},\\
&\ \ \ \ \ \ \ \ \ \ \ \ \ \ \ \ \ \ \ \ \  w_i, \mathbf{w}_{\neg i}, \mathbf{w}^{(\neg i)}, \alpha, \beta)\\
&= \int_{\phi^{(k)}} \int_{\theta^{(w_i)}} p(z_i = k, w^{(i)}_1, \ldots, w^{(i)}_{C_i}|\phi^{(k)}, \theta^{(w_i)}, w_i)\\
&\ \ \ \ \ \ \ \ \ \ \ \ \ \ \ \times  p(\phi^{(k)}, \theta^{(w_i)}|\mathbf{z}_{\neg i}, w_i, \mathbf{w}_{\neg i}, \mathbf{w}^{(\neg i)}, \alpha, \beta)\\
&= \int_{\phi^{(k)}} \int_{\theta^{(w_i)}} \theta^{(w_i)}_k  \prod_{c=1}^{C_i}\phi^{(k)}_{w^{(i)}_c} \times p(\theta^{(w_i)}|\mathbf{z}_{\neg i: w_j = w_i}, \alpha) \\
&\ \ \ \ \ \ \ \ \ \ \ \ \ \ \ \times p(\phi^{(k)}|\mathbf{z}_{\neg i}, \mathbf{w}^{(\neg i)}, \beta)\\
&= \int_{\theta^{(w_i)}} \theta^{(w_i)}_k  p(\theta^{(w_i)}|\mathbf{z}_{\neg i: w_j = w_i}, \alpha) \\
&\ \ \  \times  \int_{\phi^{(k)}}  \prod_{c=1}^{C_i}\phi^{(k)}_{w^{(i)}_c} p(\phi^{(k)}|\mathbf{z}_{\neg i}, \mathbf{w}^{(\neg i)}, \beta) \mbox{ .}\\
\end{align*}
We recognize the first integral as the mean of a Dirichlet distribution which we obtain via conjugacy:
\begin{align*}
p(\theta^{(w_i)}|\mathbf{z}_{\neg i: w_j = w_i}, \alpha) &= \mbox{Dirichlet}(\mathbf{n}_{:}^{(w_i)\neg i} + \alpha) \\
\int_{\theta^{(w_i)}} \theta^{(w_i)}_k  p(\theta^{(w_i)}|\mathbf{z}_{\neg i: w_j = w_i}, \alpha) &= \frac{n_k^{(w_i)\neg i} + \alpha_k}{\sum_{k'}n_{k'}^{(w_i)\neg i} + \alpha_{k'}}\\
&\propto n_k^{(w_i)\neg i} + \alpha_k \mbox{ .}
\end{align*}
The above can also be understood as the probability of the next ball drawn from a multivariate Polya urn model, also known as the Dirichlet-compound multinomial distribution, arising from the posterior predictive distribution of a discrete likelihood with a Dirichlet prior.  We will need the full form of such a distribution to analyze the second integral.  Once again leveraging conjugacy, we have:
\begin{align*}
&\int_{\phi^{(k)}} \prod_{c=1}^{C_i}\phi^{(k)}_{w^{(i)}_c} p(\phi^{(k)}|\mathbf{z}_{\neg i}, \mathbf{w}^{(\neg i)}, \beta)\\
&=  \int_{\phi^{(k)}}  \prod_{c=1}^{C_i}\phi^{(k)}_{w^{(i)}_c} \frac{\Gamma(\sum_{v=1}^D (n^{(k)\neg_i}_v + \beta_v))}{\prod_{v=1}^D\Gamma(n^{(k)\neg_i}_v+ \beta_v)}\prod_{v=1}^D {\phi^{(k)}_v}^{n^{(k)\neg_i}_v + \beta_v-1}\\
&=  \int_{\phi^{(k)}} \frac{\Gamma(\sum_{v=1}^D (n^{(k)\neg_i}_v+ \beta_v) )}{\prod_{v=1}^D\Gamma(n^{(k)\neg_i}_v+ \beta_v )}\prod_{v=1}^D {\phi^{(k)}_v}^{n^{(k)\neg_i}_v + \beta_v + n^{(i)}_v -1}\\
\end{align*}
\begin{align*}
&= \frac{\Gamma(\sum_{v=1}^D (n^{(k)\neg_i}_v + \beta_v) )}{\prod_{v=1}^D\Gamma(n^{(k)\neg_i}_v + \beta_v )} \frac{\prod_{v=1}^D\Gamma(n^{(k)\neg_i}_v + \beta_v + n^{(i)}_v )}{\Gamma(\sum_{v=1}^D (n^{(k)\neg_i}_v + \beta_v + n^{(i)}_v))}\\
 &\times \int_{\phi^{(k)}} \frac{\Gamma(\sum_{v=1}^D (n^{(k)\neg_i}_v + \beta_v + n^{(i)}_v) )}{\prod_{v=1}^D\Gamma(n^{(k)\neg_i}_v + \beta_v + n^{(i)}_v )}\prod_{v=1}^D {\phi^{(k)}_v}^{n^{(k)\neg_i}_v + \beta_v + n^{(i)}_v -1}\\
 &= \frac{\Gamma(\sum_{v=1}^D (n^{(k)\neg_i}_v+ \beta_v) )}{\prod_{v=1}^D\Gamma(n^{(k)\neg_i}_v+ \beta_v )} \frac{\prod_{v=1}^D\Gamma(n^{(k)\neg_i}_v + \beta_v + n^{(i)}_v )}{\Gamma(\sum_{v=1}^D (n^{(k)\neg_i}_v + \beta_v + n^{(i)}_v))} \mbox{ ,}
\end{align*}
where $n^{(i)}_v$ is the number of times that output word $v$ occurs in the $i$th context, since the final integral is over the full support of a Dirichlet distribution, which integrates to one.  Eliminating terms that aren't affected by the $z_i$ assignment, the above is
\begin{align*}
&\propto \frac{\prod_{v=1}^D\Gamma(n^{(k)\neg_i}_v + \beta_v + n^{(i)}_v )}{\Gamma(\sum_{v=1}^D (n^{(k)\neg_i}_v + \beta_v + n^{(i)}_v) )}\\
&= \frac{\prod_{v=1}^D\Big (\Gamma(n^{(k)\neg_i}_v + \beta_v) \prod_{j=0}^{n^{(i)}_v-1} (n^{(k)\neg_i}_v + \beta_v + j ) \Big )}{\Gamma(\sum_{v=1}^D (n^{(k)\neg_i}_v + \beta_v)) \prod_{j=0}^{C_i-1}(\sum_{v=1}^D (n^{(k)\neg_i}_v + \beta_v) + j )}\\
&\propto \frac{\prod_{v=1}^D \prod_{j=0}^{n^{(i)}_v-1} (n^{(k)\neg_i}_v + \beta_v + j )}{\prod_{j=0}^{C_i-1}(\sum_{v=1}^D (n^{(k)\neg_i}_v + \beta_v) + j )}\\
&= \prod_{c=1}^{C_i} \frac{n^{(k)\neg_i}_{w_c} + \beta_{w_c} + n_{w_c^{(i,c)}}}{n^{(k)\neg_i} + \sum_v \beta_v + c-1}\\
\end{align*}
where we have used the fact that $\Gamma(x + n) = (x+n-1)(x+n-2)...(x+1)x\Gamma(x)$ for any $x>0$, and integer $n\geq 1$.  We can interpret this as the probability of drawing the context words under the multivariate Polya urn model, in which the number of ``colored balls'' (word counts plus prior counts) is increased by one each time a certain color (word) is selected.  In other words, in each step, corresponding to the selection of each context word, we draw a ball from the urn, then put it back, \emph{along with another ball of the same color}.  The $n_{w_c^{(i,c)}}$ and $c-1$ terms reflect that the counts have been changed by adding these extra balls into the urn in each step.  The second to last equation shows that this process is exchangeable: it does not matter which order the balls were drawn in when determining the probability of the sequence.  Multiplying this with the term from the first integral, calculated earlier, gives us the final form of the update equation,
\begin{align*}
p(z_i &= k|\cdot )
\propto (n_k^{(w_i)\neg i} + \alpha_k) \prod_{c=1}^{C_i} \frac{n^{(k)\neg_i}_{w_c} + \beta_{w_c} + n_{w_j^{(i,c)}}}{n^{(k)\neg_i} + \sum_v \beta_v + c-1} \mbox{ .}
\end{align*}

\end{document}